\documentclass[Afour,sageh,times]{sagej}

\usepackage{moreverb,url}

\usepackage[colorlinks,bookmarksopen,bookmarksnumbered,citecolor=red,urlcolor=red]{hyperref}
\usepackage{epstopdf}

\usepackage{array}
\usepackage{wasysym}
\usepackage{bm}
\usepackage[]{units}

\newcommand\BibTeX{{\rmfamily B\kern-.05em \textsc{i\kern-.025em b}\kern-.08em T\kern-.1667em\lower.7ex\hbox{E}\kern-.125emX}}

\def\volumeyear{2019}

\begin{document}

\runninghead{Smith and Wittkopf}


\title{Dynamic Locomotion For Passive-Ankle Biped Robots And Humanoids Using Whole-Body Locomotion Control}

\author{D. Kim\affilnum{1}, S. Jorgensen\affilnum{2}, J. Lee\affilnum{3}, J. Ahn\affilnum{3}, J. Luo\affilnum{4}, and L. Sentis\affilnum{5}}

\email{dk6587@utexas.edu}
\email{stevenjj@utexas.edu}
\email{jmlee87@utexas.edu}
\email{junhyeokahn91@utexas.edu}
\email{jianwen@stanford.edu}

\affiliation{\affilnum{1}University of Texas at Austin, ASE, US \\
\affilnum{2}University of Texas at Austin, ME, US. NASA Space Technology Research Fellow (NSTRF)\\
\affilnum{3}University of Texas at Austin, ME, US\\
\affilnum{4}Stanford University, CS, US \\
\affilnum{5}University of Texas at Austin, ASE, US}
\email{lsentis@austin.utexas.edu}

\begin{abstract}
Whole-body control (WBC) is a generic task-oriented control method for feedback control of loco-manipulation behaviors in humanoid robots. The combination of WBC and model-based walking controllers has been widely utilized in various humanoid robots. However, to date, the WBC method has not been employed for unsupported passive-ankle dynamic locomotion. As such, in this paper, we devise a new WBC, dubbed whole-body locomotion controller (WBLC), that can achieve experimental dynamic walking on unsupported passive-ankle biped robots. A key aspect of WBLC is the relaxation of contact constraints such that the control commands produce reduced jerk when switching foot contacts. To achieve robust dynamic locomotion, we conduct an in-depth analysis of uncertainty for our dynamic walking algorithm called time-to-velocity-reversal (TVR) planner. The uncertainty study is fundamental as it allows us to improve the control algorithms and mechanical structure of our robot to fulfill the tolerated uncertainty. In addition, we conduct extensive experimentation for: 1) unsupported dynamic balancing (i.e. in-place stepping) with a six degree-of-freedom (DoF) biped, Mercury; 2) unsupported directional walking with Mercury; 3) walking over an irregular and slippery terrain with Mercury; and 4) in-place walking with our newly designed ten-DoF viscoelastic liquid-cooled biped, DRACO. Overall, the main contributions of this work are on: a) achieving various modalities of unsupported dynamic locomotion of passive-ankle bipeds using a WBLC controller and a TVR planner, b) conducting an uncertainty analysis to improve the mechanical structure and the controllers of Mercury, and c) devising a whole-body control strategy that reduces movement jerk during walking. 
\end{abstract}


\maketitle
	
\section{Introduction}

Passive-ankle walking has some key differences with respect to ankle actuated biped legged locomotion: 1) bipeds with passive ankles have lesser degrees-of-freedom (DoF) than ankle actuated legged robots resulting in lower mechanical complexity and lighter lower legs. 2) bipeds with passive ankles have tiny feet which lead to a small horizontal footprint of the robot. Our paper targets passive and quasi-passive ankle legged robots in leverage the above characteristics. In addition, there is a disconnect between dynamic legged locomotion methods, e.g. \cite{Rezazadeh:vk, Hartley:2017ho} and humanoid control methods, e.g. \cite{Koolen:2016ci,escande2014hierarchical, Kuindersma:2015cw}, the latter focusing on coordinating loco-manipulation behaviors. Humanoid robots like the ones used during the DARPA robotics challenges (DRC) have often employed task-oriented inverse kinematics and inverse dynamics methods coupled with control of the robots' horizontal center of mass (CoM) demonstrating versatility for whole-body behaviors \cite{Kohlbrecher:2014jx, Feng:2015ix,Johnson:2015hv, Radford:2015ca}. However, they have been practically slower and less robust to external disturbances than bipeds employing dynamic locomotion methods which do not rely on horizontal CoM control. This paper aims to explore and offer a solution to close the gap between these two lines of controls, i.e. versatile task-oriented controllers and dynamic locomotion controllers. 

There is a family of walking control methods \cite{8360164,Raibert:1984ej} that do not rely on explicit control of the horizontal CoM movement enabling passive-ankle walking and also fulfilling many of the benefits listed above. These controllers use foot placements as a control mechanism to stabilize the under-actuated horizontal CoM dynamics. At no point, they attempt to directly control the CoM instantaneous state. Instead, they calculate a control policy in which the foot location is a feedback weighted sum of the sensed CoM state. Our dynamic locomotion control policy falls into this category of controllers albeit using a particular CoM feedback gain matrix based on the concept of time-to-velocity-reversal (TVR) \cite{Kim:2014wua}. Another important dynamic locomotion control strategy relies on the concept of hybrid zero dynamics (HZD) \cite{Westervelt:2007tn}. HZD considers an orbit for dynamic locomotion and a feedback control policy that warranties asymptotic stability to the orbit \cite{Hartley:2017ho,Hereid:fs}. Although these two lines of dynamic walking controls have had an enormous impact in the legged locomotion field, they have not been extended yet to full humanoid systems. In particular, humanoid systems employing task-based whole body control strategies require closing the gap with the above dynamic locomotion methods. And this is precisely the main objective of this paper.  

The main contribution of this paper is to achieve unsupported dynamic walking of passive-ankle and full humanoid robots using the whole-body control method. To do so, we: 1) devise a new task-based whole-body locomotion controller that fulfills maximum tracking errors and significantly reduces contact jerks; 2) conduct an uncertainty analysis to improve the robot mechanics and controls; 3) integrate the whole-body control method with our dynamic locomotion planner into two experimental bipeds robots, and 4) extensively experiment with unsupported dynamic walking such as throwing balls, pushing a biped, or walking in irregular terrains.

One important improvement we have incorporated in our control scheme is to switch from joint torque control to joint position control. This low-level control change is due to the lessons we have learned regarding the overall system performance difference between low-level joint control versus torque control. Namely that joint position control used in this paper works better than a joint torque control \cite{Kim:2016jg}. Additionally, our decision to use a low level joint-level control is supported by previous studies that torque control reduces the ability to achieve a high-impedance behavior \cite{Calanca:2016kc}, which is needed for achieving dynamic biped locomotion with passive-ankle bipeds. Indeed, switching to joint position control has been a strong performance improvement to achieve the difficult experimental results. 

From the uncertainty analysis of our TVR dynamic locomotion planner, we found that to achieve stable locomotion the robot requires higher position tracking accuracy than initially expected. Our uncertainty analysis concludes that the landing foot positions need to be controlled within a 1 \unit{cm} error and the CoM state needs to be estimated within a 0.5 \unit{cm} error. Both the robot's posture control and the swing foot control require high tracking accuracy. For this reason, we remove the torque feedback in the low-level controller and instead impose a feedforward current command to compensate for whole-body inertial, Coriolis, and gravitational effects. However, this is not enough to overcome friction and stiction of the joint drivetrain. To overcome this issue, we introduce a motor position feedback controller \cite{Pratt:vt}.

Next, the low-level joint commands are computed by our proposed whole-body locomotion controller (WBLC). WBLC consists of two sequential blocks: a kinematics-level whole-body control, hereafter referred to as (KinWBC) and a dynamics-level whole-body controller (DynWBC). The first block, KinWBC, computes joint position commands as a function of  the desired operational task commands using feedback control over the robot's body posture and its foot position. 

Given these joint position commands, DynWBC computes feedforward torque commands while incorporating gravity and Coriolis forces, as well as friction cone constraints at the contact points. One key characteristic of DynWBC is the formulation of reduced jerk torque commands to handle sudden contact changes. Indeed, in our formulation, we avoid formulating contacts as hard constraints \cite{Herzog:2016ce, Saab:2013ka,Wensing:2013fm} and instead include them as a cost function. We then use the cost weights associated with the contacts to change behavior during contact transitions in a way that it significantly reduces movement jerk. For instance, when we apply heavy cost weights to the contact accelerations, we effectively emulate the effect of contact constraints. During foot detachment, we continuously reduce the contact cost weights. By doing so, we accomplish smooth transitions as the contact conditions change. An approach based on whole-body inverse dynamics has been proposed for smooth task transitions \cite{Salini:2011bw}, but has not been proposed for contact transitions like ours, neither has it been implemented in experimental platforms.

The above WBLC and joint-level position feedback controller can achieve high fidelity real-time control of bipeds and humanoid robots. For locomotion control, we employ the time-to-velocity-reversal (TVR) planner presented in \cite{Kim:2016jg}. We use the TVR planner to update foot landing locations at every step as a function of the CoM state. And we do so by planning in the middle of leg swing motions. By continuously updating the foot landing locations, bipeds accomplish dynamic walking that is robust to control errors and to external disturbances. The capability of our walking controller is extensively tested in a passive-ankle biped robot and in a quasi-passive ankle lower body humanoid robot. By relying on foot landing location commands, our control scheme is generic to various types of bipeds and therefore, we can accomplish similar walking capabilities across various robots by simply switching the robot parameters. To demonstrate the generality of our controller, we test not only two experimental bipeds but also a simulation of other humanoid robots. 

Indeed, experimental validation is a main contribution of this paper. The passive ankle biped, Mercury, is used for extensive testing of dynamic balancing, directional walking, and rough terrain walking. We also deploy the same methods to our new biped, DRACO, and accomplish dynamic walking within a few days after the robot had its joint position controllers developed. Such timely deployment showcases the robustness and versatility of the proposed control framework. 

The paper is organized as follows. Section.~\ref{sec:robot_description} introduces our robot hardware and its characteristic features. In section~\ref{sec:ctrl_scheme}, we explain the control framework consisting of the dynamic locomotion planner, whole-body locomotion controller, and the joint-level position controller. Section~\ref{sec:uncertainty_analysis} explains how the measurement noise and landing location error affect the stability of our dynamic locomotion controller, and analyzes the required accuracy for state estimation and swing foot control to asymptotically stabilize bipeds. In section~\ref{sec:implementation}, we address implementation details. Section~\ref{sec:experiment} discusses extensively experimental and simulation results. Finally, section~\ref{sec:conclusion} concludes and summarizes our works. 
\begin{figure*}
    \centering
    \includegraphics[width=2.\columnwidth]{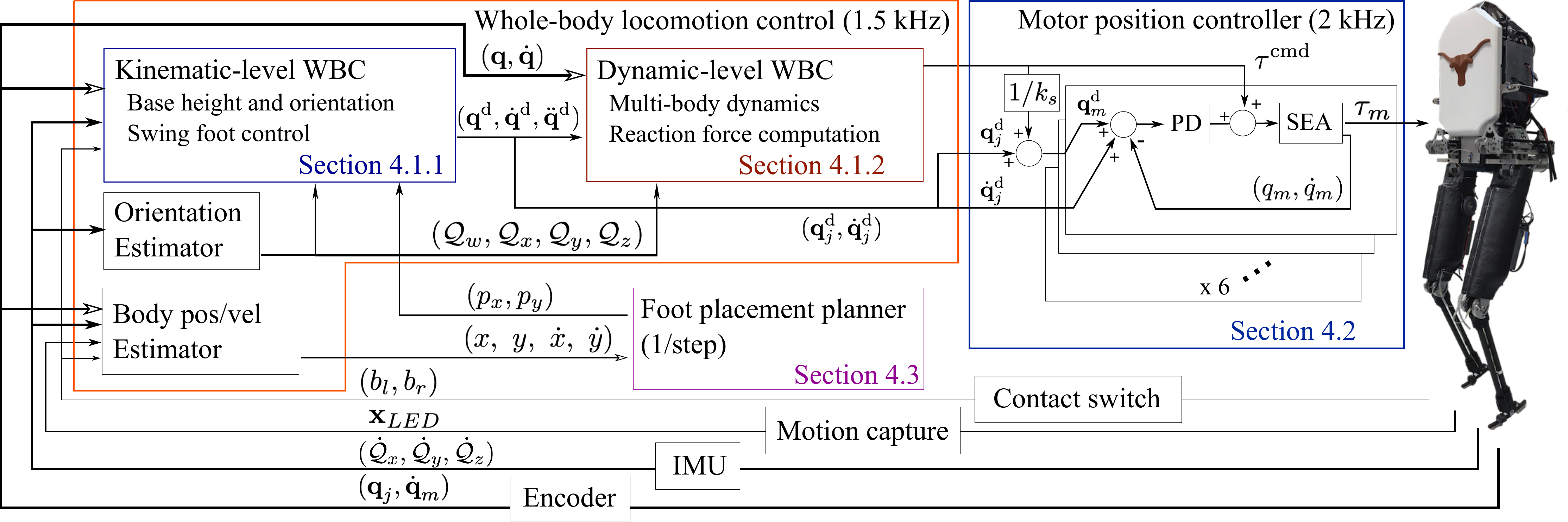}
    \caption{{\bf Diagram of the WBLC control scheme.} Our WBLC controller has a cascaded structure with three feedback loops. $b_l$ and $b_r$ are left and right foot contact signals. $\mathbf{x}_{\rm LED}$ is the position of the sensed MoCap LED markers. (1) An inverse kinematics controller computes joint positions and their first two derivatives based on desired operational space tasks. (2) An inverse dynamics whole-body controller computes contact-consistent torque commands to handle robot dynamics subject to unilateral constraints. (3) Low-level PD controllers on motor positions with feedforward currents from the computed torques are used to achieve the desired joint configurations. A TVR footstep planner plans foot landing locations, one per step at the midpoint of swing leg trajectories. Note that $\mathbf{q}_m$ are joint positions computed from the motor positions via a transmission ratio and $\mathbf{q}_j$ are joint positions measured by absolute joint encoders.}
    \label{fig:ctrl_scheme}
\end{figure*}

\section{Related Work}

ATRIAS~\cite{Rezazadeh:vk, Hartley:2017ho} is the closest example to our proposed work for this paper. It is one of the first passive-ankle biped robots that is able to dynamically balance and walk unsupported. The key difference is that our framework focuses on methods applicable to whole-body humanoid robot control applied to passive-ankle bipeds. Both our proposed dynamic locomotion planner and whole-body locomotion controller are novel and unique. In particular, our dynamic locomotion planner is based on the concept of time-to-velocity-reversal and incorporate an uncertainty analysis that drive the mechanical and control design of robots to improve their performance. In addition, our whole-body locomotion controller evolves from a long line of research on task-based whole-body humanoid robot control from our group by incorporating tools to significantly reduce contact-induced movement jerk.

There are several pioneering examples of dynamic biped locomotion: ATLAS~\cite{bostondynamics:atlas} and ASIMO~\cite{honda:asimo}. It is difficult to tell how much these robots rely on ankle actuation and foot support because of the lack of published work. It is also impossible for us to tell what kind of dynamic locomotion planners and whole-body control methods are implemented. 

Passive walking robots \cite{McGeer:1990uk,Collins:2005jx} fall in the dynamic locomotion category too. These studies shed light on the important aspects of biped locomotion, but do not provide direct application for feedback control related to our methods. On the other hand, the progress made in actuated planar biped locomotion is impressive. \cite{raibert1989dynamically, Sreenath:vb} show biped robots running and their capability to recover from disturbances on irregular terrains. However, there is an obvious gap between supported (or constrained) locomotion and unsupported walking. \cite{Raibert:1984ej} shows unsupported single leg hopping, which is a remarkable accomplishment. Besides the strong contribution in dynamic locomotion of that work, the study omitted several important aspects of unsupported biped locomotion such as body posture control, continuous interaction of the stance leg through the ground contact phases, and disturbances from the other limbs' motion, which are a focus of our paper.

\section{Locomotion Control Architecture}
\label{sec:ctrl_scheme}
Our proposed control architecture consists of three components: 1) a whole-body locomotion controller (WBLC) which coordinates joint commands based on desired operational space goal trajectories, 2) a set of joint-level feedback controllers which execute the commanded joint trajectories, and 3) a dynamic locomotion planner for passive-ankle bipeds which generates the foot landing locations based on TVR considerations. In this section, we will describe the details of these layers as well as their interaction. 

\subsection{Whole-Body Locomotion Controller (WBLC)}
Many WBCs include a robot dynamic model to compute joint torque/force commands to achieve desired operational space trajectories. If we had ideal motors with perfect gears, the computed torque commands of a WBC could be sent out as open-loop motor currents. However, excluding some special actuator designs \cite{Wensing:2017ko}, it is non-trivial to achieve the desired torque/force commands using open-loop motor currents because most actuators have high friction and stiction in their drivetrains. One established way to overcome drivetrain friction is to employ torque/force sensor feedback at the joint level. However, negative torque/force feedback control is known to reduce the maximum achievable close-loop stiffness of joint controllers \cite{Calanca:2016kc}. In addition, torque/force feedback controllers used in combination with position control are known to be more sensitive to contact disturbance and time delay. Therefore, we need a solution that addresses all of these limitations. 

Another consideration is related to the task space impedance behavior that is needed to achieve dynamic walking. Our observation is that a high impedance behavior in task space is preferred for dynamic walking because: 1) the foot landing location must be fairly accurate to stabilize the biped; 2) the swing leg must be able to overcome external disturbances; and 3) the robot's body posture needs to suppress oscillations caused by the effect of moving limbs or other disturbances. High stiffness control of robots with sizable mechanical imperfections is the only way to achieve stable passive-ankle biped walking despite making them less compliant with respect to the terrain. 

To accomplish high gain position control, we have opted to remove sensor-based torque feedback at the joint level and replace it with motor position feedback control. Our observation is that this change significantly reduces the effect of the imperfect mechanics and achieves higher position control bandwidth than using torque feedback. In addition to the joint position commands, the desired torque commands computed via WBC are incorporated as feedforward motor current commands. Thus, to combine motor position and feedforward motor current commands for dynamic locomotion, we devise a new WBC formulation that we call whole-body locomotion control (WBLC). 

WBLC is sequentially implemented with two control blocks. The first block is a kinematic-level WBC (KinWBC) that computes joint position, velocity, and acceleration commands. KinWBC does not rely on a dynamical model of the robot, instead it relies only on a kinematics model to coordinate multiple prioritized operation space tasks. The second block, called the dynamic-level WBC (DynWBC), takes the joint commands from KinWBC and computes the desired torque commands that are consistent with the robot dynamics and the changing contact constraints. The output of WBLC is therefore comprised of desired joint torque, position, and velocity commands, which are sent out to the joint-level feedback controllers. 

\subsubsection{Kinematic-level Whole-Body Controller (KinWBC) }
\label{sec:KinWBC}

We first formulate a kinematic whole-body controller to obtain joint commands given operational space commands. The basic idea is to compute incremental joint positions based on operational space position errors and add them to the current joint positions. This is done using null-space task prioritization as follows.   
\begin{align}
\label{eq:KinWBC_iter}
\Delta \mathbf{q}_1 & = \bm{J}_1^{\dagger}(\mathbf{x}_1^{des} - \mathbf{x}_1),\\
    \Delta \mathbf{q}_2 & = \Delta \mathbf{q}_1 + \bm{J}_{2|pre}^{\dagger}(\mathbf{x}_2^{\rm des} - \mathbf{x}_2 - \bm{J}_2\Delta \mathbf{q}_1), \\\nonumber
    & \quad \vdots \\\label{eq:qi}
    \Delta \mathbf{q}_i & = \Delta \mathbf{q}_{i-1} + \bm{J}_{i|pre}^{\dagger}(\mathbf{x}^{\rm des}_i - \mathbf{x}_i - \bm{J}_i\Delta \mathbf{q}_{i-1}),
\end{align}
where $\bm{J}_i$, $\mathbf{x}_i^{\rm des}$, and $\Delta \mathbf{q}_i$ are the $i$th task Jacobian, a desired position of the $i$th task, and the change of joint configuration related to the $i$ th task iteration. The $\{\cdot \}^{\dagger}$ denotes an SVD-based pseudo-inverse operator in which small singular values are set to 0. Note that there is no feedback gain terms in this formulation, which can be interpreted as gains being equal to unity. In addition, the prioritized Jacobians take the form: 
\begin{align}
    &\bm{J}_{i|pre}  = \bm{J}_i\bm{N}_{i-1}, \\
    &\bm{N}_{i-1} = \bm{N}_{1|0} \cdots \bm{N}_{i-1|i-2}, \\
    &\bm{N}_{i-1|i-2}  = \bm{I} - \bm{J}_{i-1|pre}^{\dagger}\bm{J}_{i-1|pre}, \\
    &\bm{N}_{0} = \bm{I}.
\end{align}
Then, the joint position commands can be found with 
\begin{equation}\label{eq:KinWBC_q}
     \mathbf{q}^{\rm d} = \mathbf{q} + \Delta \mathbf{q},
\end{equation}
where $\Delta \mathbf{q}$ is joint increment computed in the $i$th task in Eq.~\eqref{eq:qi}.
In addition, the joint velocity and acceleration for every task iteration can be computed as,
\begin{align}
\label{eq:KinWBC_qdot}
\dot{\mathbf{q}}^{\rm d}_i &= \dot{\mathbf{q}}^{\rm d}_{i-1} + \bm{J}_{i|pre}^{\dagger}\left( \dot{\mathbf{x}}^{\rm des} - \bm{J}_i\dot{\mathbf{q}}_{i-1}^{\rm d} \right), \\[2mm]
\ddot{\mathbf{q}}^{\rm d}_i &=\ddot{\mathbf{q}}^{\rm d}_{i-1} + \bm{J}_{i|pre}^{\dagger}\left(\ddot{\mathbf{x}}^{\rm des} - \dot{\bm{J}}_i\dot{\mathbf{q}} - \bm{J}_i \ddot{\mathbf{q}}^{\rm d}_{i-1} \right).
\end{align}
Finally, the joint commands, $\mathbf{q}^{\rm d}$, $\dot{\mathbf{q}}^{\rm d}$, and $\ddot{\mathbf{q}}^{\rm d}$ are sent out to the block, DynWBC. We note that $\mathbf{q}$ is the full configuration of the robot containing both floating base and actuated joints.

\subsubsection{Dynamic-level Whole-Body Controller (DynWBC) }
\label{sec:DynWBC}
Given joint position, velocity, and acceleration commands from the KinWBC, the DynWBC computes torque commands while considering the robot dynamic model and various constraints. The optimization algorithm to compute torque commands in DynWBC is as follows: 
%
\begin{align}\label{eq:qp_cost_wblc}
& \min_{\mathbf{F}_r, \ddot{\mathbf{x}}_c, \bm{\delta}_{\ddot{\mathbf{q}}}}
& &  \mathbf{F}^{\top}_r \bm{W}_{r} \mathbf{F}_r + \ddot{\mathbf{x}}_c^{\top} \bm{W}_c \ddot{\mathbf{x}}_c + \bm{\delta}_{\ddot{\mathbf{q}}}^{\top} \bm{W}_{\ddot{\mathbf{q}}} \bm{\delta}_{\ddot{\mathbf{q}}}
\\
& \text{s.t.} 
\label{eq:wblc_cone_const}
& &\bm{U}\mathbf{F}_r \geq \mathbf{0},   \\[1mm] 
\label{eq:wblc_frz_bound}
& & & \bm{S}\mathbf{F}_{r} \leq \mathbf{F}_{r,z}^{\rm max}, \\[1mm] 
\label{eq:xc_acc_wblc}
& & & \ddot{\mathbf{x}}_c = \bm{J}_c\ddot{\mathbf{q}} + \dot{\bm{J}}_c\dot{\mathbf{q}},\\[1mm]
\label{eq:wblc_multibody_dyn}
& & & \bm{A}\ddot{\mathbf{q}}   + \mathbf{b} + \mathbf{g}  = \begin{pmatrix} \mathbf{0}_{6 \times 1}  \\ \bm{\tau}^{\rm cmd} \end{pmatrix} + \bm{J}^{\top}_c \mathbf{F}_r,\\[1mm]
\label{eq:wblc_qddot}
& & & \ddot{\mathbf{q}}  = \ddot{\mathbf{q}}^{\rm cmd} + \bm{\delta}_{\ddot{\mathbf{q}}},\\[1mm]
\label{eq:wblc_qddot_cmd}
& & & \ddot{\mathbf{q}}^{\rm cmd} = \ddot{\mathbf{q}}^{\rm d} + k_d (\dot{\mathbf{q}}^{\rm d} - \dot{\mathbf{q}}) +  k_p (\mathbf{q}^{\rm d} - \mathbf{q}), \\[1mm]
\label{eq:wblc_torque_bound}
& & & \bm{\tau}_{\rm min} \leq \bm{\tau}^{\rm cmd} \leq \bm{\tau}_{\rm max}.
\end{align}
%
In turn, these computed torque commands are sent out as feedforward motor current commands to the joint-level controllers. One key difference with other QP formulations for whole-body control is that we do not use the null space operators of the contact constraints nor do we use a null velocity or acceleration assumption to describe the surface contacts of the robot with the ground. Instead, contact interactions are addressed with contact acceleration terms in the cost function regulated with weighting matrices that effectively model the changes in the contact state. This new term is particularly important since traditional modeling of contacts as hard constraints causes torque command discontinuities due to sudden contact switches. As such, we call our formulation reduced ``jerk'' whole-body control. We note that our formulation is the first attempt that we know of to use WBC for unsupported passive-ankle dynamic locomotion in experimental bipeds. Contact changes in passive-ankle biped locomotion are far more sudden than changes on robots that control the horizontal CoM movement. Our proposed formulation emerges from extensive experimentation and comparison between QP-based WBC formulations using hard contact constraints versus soft constraints as proposed. We report that the above formulation has empirically shown to produce rapidly changing but smooth torque commands than using WBCs with hard constraints.  

To achieve smooth contact switching, the contact Jacobian employed above includes both the robot's feet contacts even if one of them is not currently in contact. As mentioned above, we never set foot contact accelerations to be zero even if they are in contact. Instead, we penalize foot accelerations in the cost function depending on whether they are in contact or not using the weight $\bm{W}_c$. When a foot is in contact, we increase the values of $\bm{W}_c$ for the block corresponding to the contact. Similarly, we reduce the values of the weights when the foot is removed from the contact. At the same time, we increase the weight $\bm{W}_r$ for the swing foot and reduce the upper bound of the reaction force $\mathbf{F}_{r,z}^{\rm max}$. In essence, by smoothly changing the upper bounds, $\mathbf{F}_{r,z}^{\rm max}$, and weights, $\bm{W}_r$ and $\bm{W}_c$, we practically achieve jerk-free walking motions. The concrete description for the weights and bounds used in our experiments are explained in Section \ref{sec:WBLC_setting}.

In the above algorithm, $\bm{U}$ computes normal and friction cone forces as described in \cite{bouyarmane2018}, and $\mathbf{F}_r$ represents contact reaction forces. Eq.~\eqref{eq:wblc_frz_bound} introduces the upper bounds on the normal reaction forces to facilitate smooth contact transitions. As mentioned before, this upper bound is selected to decrease when the foot contacts detach from the ground and increase again when the foot makes contact. 

Eq.~\eqref{eq:wblc_multibody_dyn} models the full-body dynamics of the robot including the reaction forces. $\bm{A}$, $\mathbf{b}$, and $\mathbf{g}$ are the generalized inertia, Coriolis, and gravitational forces, respectively. The diagonal terms of the inertia matrix include the rotor inertia of each actuator in addition to the linkage inertia. The rotor inertia is an important inclusion to achieve good performance. Eq.~\eqref{eq:wblc_qddot} shows the relaxation of the joint commands, $\ddot{\mathbf{q}}^{\rm cmd}$, by the term, $\bm{\delta}_{\ddot{\mathbf{q}}}$. We include this relaxation because of two reasons. First, the KinWBC specifies virtual joint acceleration which cannot be perfectly attained. Second, the torque limit on the above optimization can prevent achieving the desired joint acceleration. Eq.~\eqref{eq:wblc_qddot_cmd} shows how the KinWBC's joint commands are used to find desired acceleration commands. Here, $\mathbf{q}^{\rm d}$, $\dot{\mathbf{q}}^{\rm d}$, and $\ddot{\mathbf{q}}^{\rm d}$ are the computed commands from KinWBC. Eq.~\eqref{eq:wblc_torque_bound} represents torque limits.

\subsection{Joint-Level Controller}
\label{sec:joint_ctrl}
Each actuated joint has an embedded control board that we use to implement the motor position PD control with feedforward torque inputs: 
\begin{equation}
    \tau_m = \tau^{\rm cmd} + 
    k_p \left(q_m^{\rm d} - q_m \right) +
    k_d \left(\dot{q}^{\rm d}_j - \dot{q}_m \right),
\end{equation}
where $\tau_m$ and $\tau^{\rm cmd}$ are the desired motor torque and computed torque command, the latter is obtained from Eq.~\eqref{eq:wblc_multibody_dyn} in the optimization problem. Thus, $\tau^{\rm cmd}$ acts as the feedforward control input. 
$\dot{q}^{\rm d}_j$ is the desired joint velocity computed from the KinWBC. It is obtained by applying the iterative algorithm in Eq.~\eqref{eq:KinWBC_qdot}.
$q_m^{\rm d}$ is a desired motor position command and is computed using the following formula,
\begin{equation}
    q_m^{\rm d} = q_j^{\rm d} + \frac{\tau^{\rm cmd}}{k_{s}},
\end{equation}
where $k_s$ is the spring constant of each SEA joint. $q_j^d$ is obtained via the iterative algorithm shown in Eq.~\eqref{eq:KinWBC_iter} $\sim$ \eqref{eq:KinWBC_q}. We incorporate this spring deflection consideration because the computation of joint positions from motor positions, $q_m$, considers only the transmission ratio, $N$, but the spring deflection is ignored in the computation. 

\subsection{Time-to-Velocity-Reversal (TVR) Planner}
At every step, a TVR planner computes foot placements as a function of the CoM state, i.e. its position and velocity. This is done around the middle of the swing foot motion. Our TVR planner operates with the principle of reversing the CoM velocity every step and it can be shown that the CoM movement is asymptotically stable. The original method was presented in our previous paper \cite{Kim:2016jg}. In this paper, we use a simplified version of TVR which considers a constant CoM height. This consideration has been beneficial on various experimental results across multiple biped robotics platforms explored in this paper. In Appendix 2 we explain the difference of our planner and the ones proposed by \cite{Raibert:1984ej}, \cite{Koolen:2012cta}, and \cite{Rezazadeh:vk}.

\section{Uncertainty Analysis Of The Planner}
\label{sec:uncertainty_analysis}

One of the biggest challenges in unsupported passive-ankle dynamic locomotion is to determine what control accuracy is needed to effectively stabilize a biped. Given that a passive-ankle biped robot cannot use ankle torques to control the robot's CoM movement, foot position accuracy, state estimation, and other related considerations become much more important in achieving the desired dynamic behavior. For instance, the CoM dynamics emerging from passive-ankle behavior evolves exponentially with time, pointing out the need to determine the tolerable foot position and body estimation errors. In this section, we develop the tools to explicitly quantify the required accuracy to achieve asymptotically stable passive-ankle dynamic locomotion. 

As previously mentioned, our TVR locomotion planner observes the CoM position and velocity states and computes a foot landing location. For our analysis and experimentation, we enforce a constant CoM height constraint. Our reliance on linear inverted pendulum (LIP) model enables a straightforward uncertainty analysis given noisy CoM state observations and landing location errors under kinematic constraints.

\subsection{Formulation of the Planner}

Our TVR planner relies on the LIP model:
\begin{equation} \label{eq:lip_acceleration}
\ddot{x} = \frac{g}{h}(x -p),
\end{equation}
where $g$ is the gravitational acceleration, $h$ is the constant CoM height value, and $p$ is the foot landing location which acts as a stabilizing input for reversing the CoM dynamics at every step. More concretely, the TVR planner aims to reverse the CoM velocity after a set time duration $t'$ by computing a new stance foot location, $p$. Note that Eq.~\eqref{eq:lip_acceleration} is linear so it has an exact solution for the CoM state, $\mathbf{x}(t)$. Thus, for a given $p$, the CoM state after a desired swing time $T$ can be described as a discrete system where $k$ corresponds to the $k$-th walking step of the robot:
\begin{align}
&\mathbf{x}_{k+1} = \bm{A} \mathbf{x}_k + \bm{B}p_k, \label{eq:lip_dynamics} \\[2mm]
&\bm{A} =\begin{bmatrix}
 \cosh(\omega T) & \omega^{-1} \sinh(\omega T) \\
 \omega \sinh(\omega T) & \cosh(\omega T)
 \end{bmatrix},\\
&\bm{B} = \begin{bmatrix}
1-\cosh(\omega T) \\ 
-\omega  \sinh(\omega T )
 \end{bmatrix},
\end{align}
where $\omega = \sqrt{g/h}$. The system above can be straightforwardly obtained by applying known second order linear ODE techniques to Eq.~\eqref{eq:lip_acceleration}. Next, let $p_k$ correspond to the foot location of the $k$-th step in a sequence of steps. Our TVR planner is based on the objective of finding a $p_k$ which reverse the CoM velocity at every step. Letting the velocity component (bottom row) of Eq.~\eqref{eq:lip_dynamics} be zero after the desired reversal time, $t' < T$, results into the quality, 
\begin{equation}
0 = \begin{bmatrix}\omega \sinh(\omega t')& \cosh(\omega t')\end{bmatrix} \mathbf{x}_k - \omega \sinh(\omega t') p_{k}. \label{eq:zero_vel_stance_foot}
\end{equation}
Solving for $p_k$ in the above equation will result in the foot landing location policy that reverses CoM velocity after $t'$. With the CoM velocity being reversed after every step, an additional bias term, $\kappa$, is added to steer the robot toward the origin. Further details about $\kappa$ can be found in \cite{Kim:2016jg}. Solving for Eq.~\eqref{eq:zero_vel_stance_foot} and including the additional $\kappa$ term, we get
\begin{equation}\label{eq:tvr_step_eq}
p_{k} = \begin{bmatrix}1 & \omega^{-1}\coth(\omega t')\end{bmatrix} \mathbf{x}_k +  \begin{bmatrix}
\kappa & 0
\end{bmatrix} \mathbf{x}_k.
\end{equation}
Incorporating the above feedback policy into Eq.~\eqref{eq:lip_dynamics}, we get the closed loop dynamics, 
\begin{align}\label{eq:tvr_closed_form}
&\mathbf{x}_{k+1} = (\bm{A} + \bm{B}\bm{K})\mathbf{x}_k,\\[2mm]
\label{eq:k_planner}&\bm{K} = \begin{bmatrix}
(1+\kappa) & \omega^{-1}\coth(\omega t')
\end{bmatrix}.
\end{align}

Notice that the control policy in Eq.~\eqref{eq:tvr_closed_form} has a simple PD control form; therefore, applying standard linear stability methods for PD control, the planner parameters, ($\kappa$, $t'$), can be tuned to achieve magnitudes such that the closed loop eigenvalues of $\bm{A}+\bm{B}\bm{K}$ are smaller than 1. In our case, we chose eigenvalues with magnitude equal to 0.8. Since our desired behavior is to take multiple small steps toward a desired reference position rather than a single big step, the eigenvalue magnitudes are intentionally set to be close to one rather than zero. The resulting motion (simulated numerically) in Fig.~\ref{fig:phase_plot_analysis}(a), shows the asymptotically converging trajectories in the phase plot. 

\begin{table}
\centering
\begin{tabular}{>{\centering}m{0.42\columnwidth} %
                >{\centering}m{0.42\columnwidth} %
                @{}m{0pt}@{}}
\specialrule{1.5pt}{1pt}{1pt}
{$\begin{bmatrix} t'_x & t'_y \end{bmatrix}$ }
& {$\begin{bmatrix} \kappa_x & \kappa_y \end{bmatrix}$ }
&\\[1mm] 
\hline
\hline 
[0.2, 0.2] & [0.16, 0.16]
&
\\[1.5mm]
\hline
\end{tabular}
\caption{{\bf Planner Parameter.} Each parameter has $x$ and $y$ components. We use the same value for both directions.}
\label{tb:planner_param}
\vspace{-1.5mm}
\end{table}

\subsection{Uncertainty Analysis}

During experimental walking tests, we observed notable body position and landing location errors due to the deflection of the mechanical linkages of the robot. We note that in our attempt to make Mercury a light-weight robot, we designed body and leg structures made of thin aluminum pieces and carbon fiber structures. In particular, the lower and upper legs of Mercury are constructed using carbon fibers without further rigid support. In addition, the abduction and flexion hip joints contain drivetrains made out of thin aluminum with pin joints that deflect when a contact occurs. Rather than focusing on the effect of these existing mechanical deformations, we decided to focus on the maximum errors that our dynamic locomotion controller can tolerate. After we found the maximum tolerances, we went back to the robot's mechanical design and replaced hip joints and the leg linkages to be significantly more rigid in order to fulfill the maximum tolerances. Therefore, our uncertainty analysis has been fundamental to drive the new mechanical structure on the original biped hardware to achieve the desired performance. 

To quantify the acceptable errors for our TVR planner, we perform here an analysis of stability borrowing ideas from robust control \cite{Bahnasawi:1989ft}. We apply some assumptions to simplify our analysis: 
\begin{enumerate}
\item The robot's step size is limited to 0.5 \unit{m} based on an approximated leg kinematic limits.
\item State-dependent errors are ignored. 
\end{enumerate}
For our analysis, we model foot landing location errors (presumably resulting from mechanical deflection and limited control bandwidth) with a scalar term, $\eta$. On the other hand, we model CoM state estimation errors as a vector of position and velocity errors, $\bm{\delta}$. Based on these error variables, we extend the dynamics of Eq.~\ref{eq:lip_dynamics} to be 

\begin{equation} \label{eq:pipm_onestep}
\begin{split}
\mathbf{x}_{k+1} &= \bm{A}\mathbf{x}_k + \bm{B}(p_{k} + \eta), \\
p_k &= \bm{K}(\mathbf{x}_k + \bm{\delta}).
\end{split}
\end{equation}
In order to provide design specifications to improve the robot mechanics, controllers, and estimation processes, we choose arbitrary bounds such that
\begin{equation}
||\bm{\delta}|| \leq \delta_{M}, \quad ||\eta|| \leq \eta_{M}.
\end{equation}
Once again, we use the proposed uncertainty analysis to determine the maximum tolerance bounds $\delta_M$ and $\eta_M$, providing design specifications. 
Since the velocity of the state resulting from our TVR planner changes sign after every step, typical convergence analysis regards this effect as an oscillatory behavior despite the fact that the absolute value of the CoM state, $\mathbf{x}$, effectively decreases over time. 
To remedy this, we perform a convergence analysis after two steps instead of a single step. Therefore, given an initial state, $\mathbf{x}$, after two steps, the new state, $\mathbf{x}''$, is obtained by applying Eq.~\eqref{eq:pipm_onestep} twice,
\begin{equation} \label{eq:pipm_twostep}
\begin{split}
&\mathbf{x}'' = \bm{A}^2\mathbf{x} + \bm{A}\bm{B}(p + \eta) + \bm{B}(p' + \eta'), \\
&p = \bm{K}\left(\mathbf{x} + \bm{\delta}\right),\\
&p' = \bm{K}\left(\mathbf{x}' + \bm{\delta}'\right),
\end{split} 
\end{equation}
where $()$, $()'$, and $()''$ represent the $k$th, $(k+1)$th and $(k+2)$th step respectively. The main idea is to find the region in $\mathbf{x}$ for which a Lyapunov function decreases value after two steps subject to the maximum errors, $\delta_M$ and $\eta_M$:
\begin{align}
\Delta V &= \mathbf{x}''^{\top}\bm{P}\mathbf{x}'' - \mathbf{x}^{\top}\bm{P}\mathbf{x} \leq 0.
\end{align}
Substituting Eq.~\eqref{eq:pipm_twostep}, arranging the terms, and setting the upper bound $\Delta V$, it can be shown that 
\begin{align}
\Delta V  &= \mathbf{x}^{\top}(\bm{A}_{cc}^{\top}\bm{P}\bm{A}_{cc}-\bm{P})\mathbf{x} + 2\bm{\zeta}^{\top}\bm{P}\bm{A}_{cc}\mathbf{x} + \bm{\zeta}^{\top}\bm{P}\bm{\zeta} \nonumber \\ \label{eq:delta_V}
 &\leq -a||\mathbf{x}||^2 + 2b||\mathbf{x}|| + c \\
 &\leq 0, \nonumber
\end{align}
where,
\begin{align}
&\bm{A}_{c} = \bm{A} -\bm{B}\bm{K}\\
&\bm{A}_{cc} = \bm{A}_c^2\\
\label{eq:zeta}
&\bm{\zeta} =\bm{A}_c\bm{B}\bm{K}\bm{\delta} + \bm{B}\bm{K}\bm{\delta}'+ \bm{A}_c\bm{B}\eta + \bm{B}\eta'\\
&a= -\lambda_M\left(\bm{A}_{cc}^{\top}\bm{P}\bm{A}_{cc}-\bm{P}\right),\\
\nonumber
&b = \delta_M\left( ||\bm{A}_{cc}^{\top}\bm{P}\bm{A}_c\bm{B}\bm{K}|| + ||\bm{A}_{cc}^{\top}\bm{P}\bm{B}\bm{K}||\right) + \\\label{eq:b}
& \quad \quad \quad \quad \quad \eta_M\left( ||\bm{A}_{cc}^{\top}\bm{P}\bm{A}_c\bm{B}|| + ||\bm{A}_{cc}^{\top}\bm{P}\bm{B}||\right),\\
&c = g(\bm{\zeta}^{\top}\bm{P}\bm{\zeta}).
\end{align}

\begin{figure}
\centering
\includegraphics[width=\columnwidth]{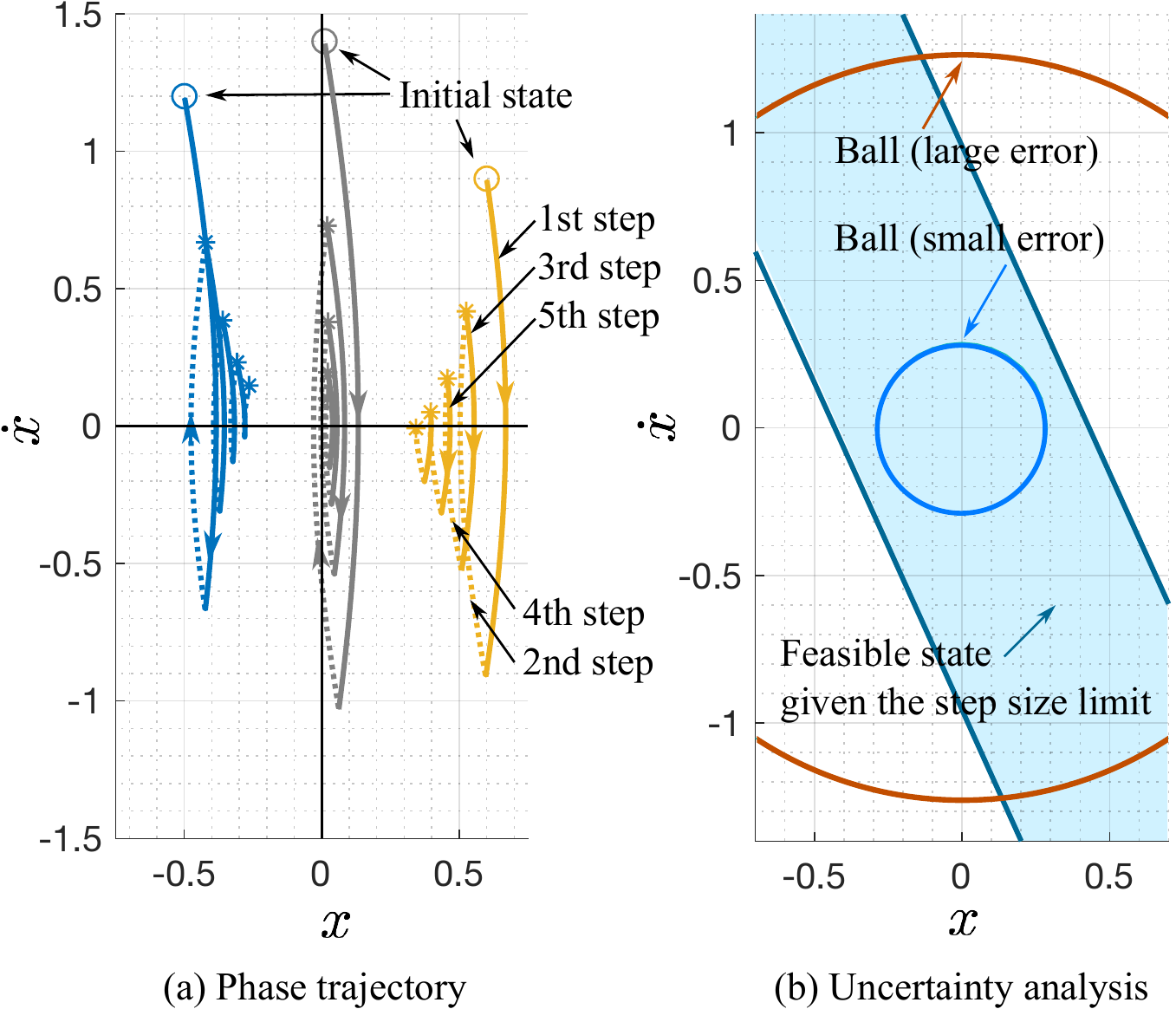}
\caption{{\bf Phase Plot and Uncertainty Analysis.} In (a), the phase trajectories of 8 steps from three initial states converging toward the origin are shown. In (b), the region of uncertainty is encircled by balls. States within the blue ball and outside of the ball radius are kinematically feasible and asymptotically stable respectively. The ball radius increases when the system has large errors in state observation and control input.}
\label{fig:phase_plot_analysis}
\end{figure}
Notice that the upper bounds defined by $a$, $b$, and $c$ have a quadratic form which allows us to find easily a solution of the Euclidean norm of the CoM state. $||\cdot||$ is the $l^2$-norm, $\lambda_{M}(\cdot)$ denotes the maximum eigenvalue of its matrix arguments, and $g(\bm{\zeta}^{\top}\bm{P}\bm{\zeta})$ is the sum of the $l^2$-norm of every term in $\bm{\zeta}^{\top}\bm{P}\bm{\zeta}$ similar to $b$. The definition of $g$ is pushed down to Appendix 3 due to the length of the expression. Note that $a$ is positive if the planner parameters are tuned such that the LIP  behavior is stable. Solving for $-a ||\mathbf{x}||^2 + 2b||\mathbf{x}||+c \leq 0$, we get the uncertainty ball region,
\begin{equation}\label{eq:radius_ball}
B_r = \left\{ \mathbf{x} \ \middle| ||\mathbf{x}|| \leq \frac{b + \sqrt{b^2+ac}}{a} \right\}.
\end{equation}
The above ball defines the region of states for which we cannot warranty asymptotic stability. And conversely, the region of states outside of the ball, $\mathbf{x} \not\in B_r$, corresponds to asymptotically stable states. Note that a smaller ball means a larger stability region, and if the errors $\eta$ and $\delta$ are zero, the ball would have zero radius and any state would be asymptotically stable. However, because of mechanical deflection, limited control bandwidth, and estimation errors, $b$ and $c$ are non-zero. 

By substituting the planner's parameters from Table~\ref{tb:planner_param} into the above equation, we can quantify and analyze the effect of the errors mentioned above. Fig.~\ref{fig:phase_plot_analysis}(b) shows the CoM phase space plot. Take Eq.~\eqref{eq:tvr_step_eq} and write it in the simple form,
\begin{equation}\label{eq:tvr_simple_form}
    p = k_p x + k_d \dot{x}.
\end{equation}
As we said, this equation corresponds to the foot landing location control policy to stabilize a biped robot. We also mention that the maximum step size for our robot, Mercury, is $-0.5\unit{m}<p<0.5\unit{m}$. If we apply these kinematic limits to the above foot control policy, we obtain a pair of lines in the phase plane which define the area of feasible CoM states given foot kinematic limits. This area is highlighted in light blue color in our phase plot. To be clear, the light blue colored area defines the state for which the robot can recover within a single walking step without violating kinematic limits. 

Next let us consider the uncertainty region defined by Eq.~\eqref{eq:radius_ball}. Notice that the terms $b$ and $c$ depend on the uncertainty errors. For example, if we have a maximum foot landing error of 0.045 \unit{m} and a maximum state estimation error of 0.03 \unit{m}, then $\eta_M = 0.045\unit{m}$ and $\delta_M = 0.03\unit{m}$. If we plug these values in Eq.~\eqref{eq:b} $\sim$ \eqref{eq:radius_ball}, we get the orange ball shown in Fig.~\ref{fig:phase_plot_analysis} (b). The inside of this ball represents states for which we cannot warranty asymptotic stability. The problem is that the orange uncertainty region include states outside of the feasible CoM state, the light blue region. This means that the actual CoM could have a value for which the robot cannot recover because it requires foot steps outside of the robot's kinematic limits. As we mentioned before, our biped robot, Mercury, underwent significant mechanical, control, and sensing improvements to remedy this problem. The errors represented by the orange ball are close to what we have observed in our walking tests before we upgraded Mercury. After making hardware and control improvement, we reduced the errors to $\eta_M = 0.01\unit{m}$ and $\delta_M = 0.007\unit{m}$. 

In particular, to reduce $\delta_M$, we employed a tactical IMU (STIM-300) and MoCap data from a phase space motion capture system providing a body velocity estimation resolution of 0.005\unit{m/s} and a body position accuracy of 0.005\unit{m}. The blue ball in Fig.~\ref{fig:phase_plot_analysis}(b) represents the new uncertainty region given by this significant improvements. We can now see that the blue ball is completely contained within the light blue region. This means that although we do not know where the CoM state is located inside the blue ball, we know that whatever the state is, it is within the feasible CoM state region, and therefore, the foot control policy will find stabilizing foot locations. 

\section{Mercury Experimental Robot}
\label{sec:robot_description}
The methods described in this paper have been extensively tested in two biped platforms. Most experiments are performed in our biped robot, Mercury, which we describe here. An additional experiment is performed in a new biped, called DRACO, which is described in \cite{JunhyeokDRACO}. Mercury has six actuators which control the hip abduction/adduction, flexion/extension, and knee flexion/extension joints. Mercury uses series-elastic actuators (SEAs), which incorporate a spring between the drivetrains and the joint outputs. The springs protect the drivetrains from external impacts and are used for estimating torque outputs at the joints. Additionally, Mercury went through significant hardware upgrades from our previous robot, Hume \cite{Kim:2016jg}. In this section, we provide an overview of our system and discuss the upgrade. We also explain similarities and differences with respect to other humanoid robots in terms of mass distribution. 

\begin{figure}
    \centering
    \includegraphics[width=0.9\columnwidth]{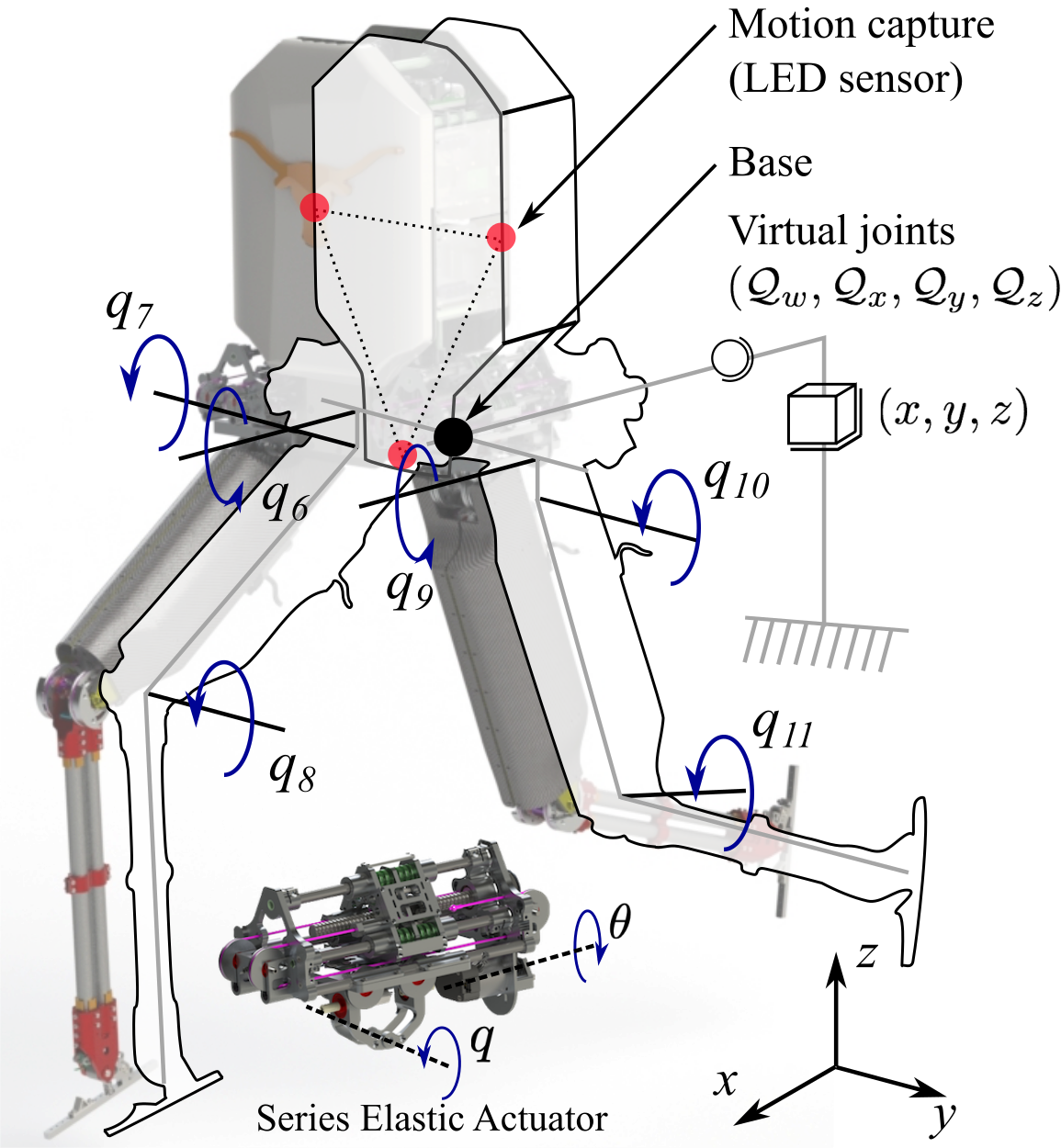}
    \caption{{\bf Configuration of Mercury.} To represent the floating base dynamics, we connect virtual joints at the base of Mercury. The virtual joints consist of three prismatic joints and a ball joint which is expressed as a quaternion. Each leg has an actuated abduction/adduction $(q_6, q_9)$, hip flexion/extension $(q_7, q_{10})$, and knee flexion/extension $(q_8, q_{11})$ joints. Lastly, three LED sensors are attached on the front of the robot's body to estimate the velocity of its physical base.}
    \label{fig:configuration}
\end{figure}
\begin{figure*}
\centering
\includegraphics[width=2.\columnwidth]{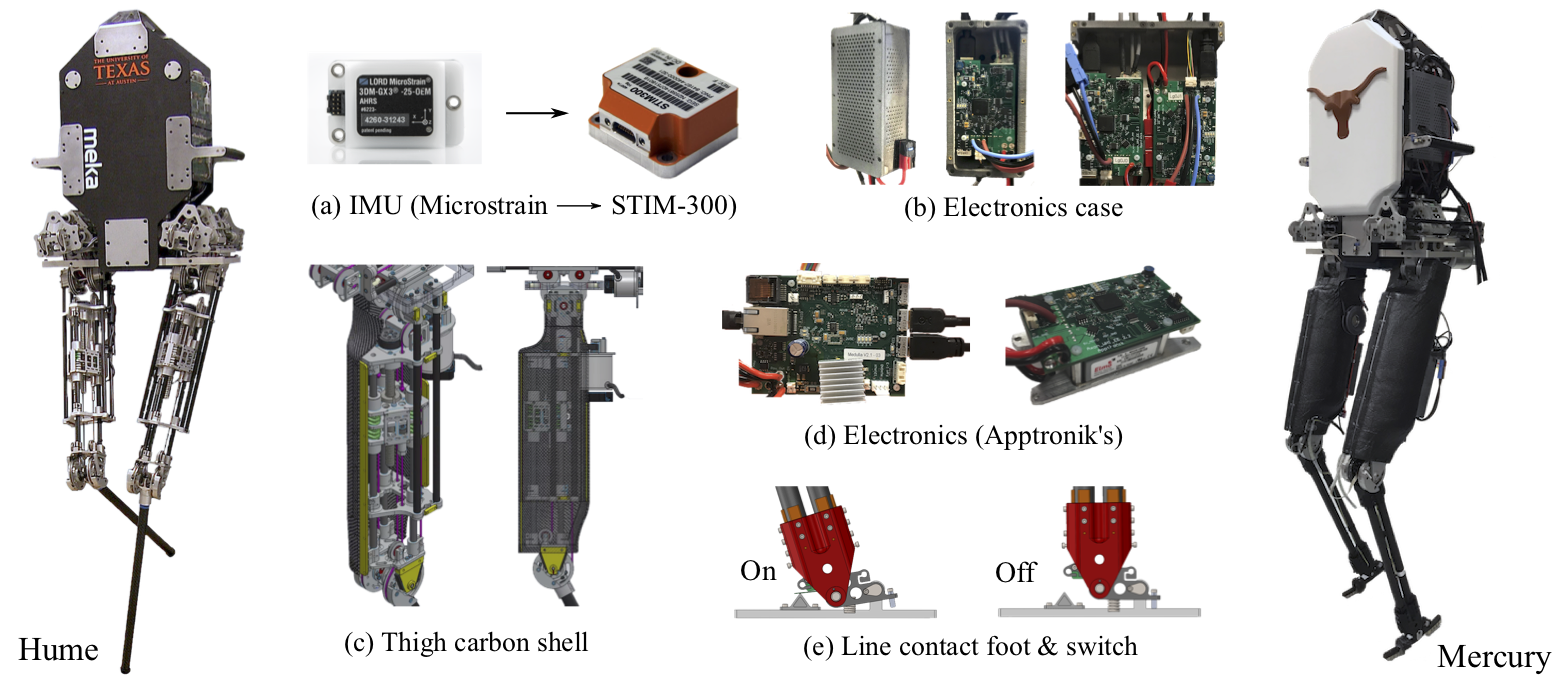}
\caption{{\bf Hardware upgrades of Mercury.} (a) The IMU was upgraded to the Sensonor's STIM-300, which has low angular velocity drift and bias, enabling accurate orientation estimates even with simple forward integration. (b) The on-board electronics have been installed with cases to secure the electric cables in place. Keeping the cables in place significantly reduces loses of connections and cable damage during robot operations. (c) Carbon fiber cases were installed on Mercury's thighs to increase structural stiffness. (d) All of the embedded electronics were replaced with Apptronik's Medulla and Axon boards that come with a variety of low-level controllers for SEAs. (e) Spring-loaded passive-ankles with limit switches were also added to limit the uncontrollable yaw body rotation and detect ground contacts.}
\label{fig:hardware_upgrade}
\end{figure*}
\subsection{Robot Configuration}
Fig.~\ref{fig:configuration} shows the sensing system and configuration of Mercury. Mercury's configuration starts from a set of virtual joints fixed to ground, representing the floating base dynamics of the robot. The end joint of the 6-DoF virtual joint is attached to the physical base of Mercury. The orientation of the virtual ball joint is represented by a quaternion and its angular velocity is represented by the space so(3) with respect to the local base frame. The actuated joints start from the right hip abduction/adduction and goes down to the hip flexion/extension, and knee flexion/extension joints. Then, the joint labels continue on to the left leg starting also at the hip joint. Three LED sensors are attached to the front of robot's body frame to estimate the robot's linear velocity and its global position via MoCap. In addition, we also estimate the relative robot position using joint encoder data with respect to the stance foot. This last sensing procedure is partially used to control foot landing locations, and therefore, the reference frame changes every time the robot switches contact.  

Mercury's SEA actuators were built in 2011 by Meka, each having three encoders to measure joint position, spring deflection, and motor position. An absolute position encoder is used to measure the joint output position, $q_j$, while a low-noise quadrature encoder measures motor position, $\theta$. Joint position and joint velocity sensing can be done either using the absolute encoder or via applying a transmission ratio transformation on the motor's quadrature encoder data ($q_m$). In our experiment, we use the absolute encoders to obtain joint positions and motor quadrature encoders to obtain joint velocities. The transmission ratio of all Mercury's joints has a constant value except for the abduction/adduction joints which are non-constant. The constant ratio occurs for transmissions consisting of a pulley mechanism with constant radius. On the other hand, the hip abduction/adduction joints consist of a spring cage directly connected to the joints which results on a change of the moment arm. To account for this change, we use a look-up table mapping the moment arm length with respect to the joint position.

\subsection{Hardware Upgrades}
The original biped, Hume, was mostly built in 2011 by Meka as a custom robot for our laboratory. It had several limitations that made dynamic locomotion difficult. It had a low-performance IMU which made it difficult to control the robot's body orientation. Hume's legs were not strong enough causing buckling of the structure when supporting the robot's body mass. Because of this structural buckling, the estimated foot positions obtained from the joint encoders was off by 5 \unit{cm} from their actual positions. We estimated this error by comparing the joint encoder data with the MoCap system data. Hume terminated its legs with cylindrical cups that would make contact with the ground. These cups had a extremely small contact surface with the ground. During walking, Hume suffered from significant vertical rotation, i.e. yaw rotation, due to the minimal contact of its supporting foot with the ground. All of these problems, i.e. structural buckling, poor IMU sensor, and small contact surfaces, prevented Hume from accomplishing stable walking. Therefore, for the proposed work, we have significantly upgraded the robot in all of these respects and changed its name to Mercury. 

To improve on state estimation, we upgraded the original IMU, a Microstrain 3DM-GX3-25-OEM, to a tactical one, a STIM-300 (Fig.~\ref{fig:hardware_upgrade}(a)). Both IMUs are MEMS-based but the bias instability of the tactical IMU is only 0.0087 \unit[per-mode=symbol]{rad/h}. Such low-bias noise allows us to estimate the robot's body orientation by simply integrating over the angular velocity from the initial orientation. Another problem we were facing with our original biped is the aging electronics, originally built by Meka in 2011. For this reason, all control boards (Fig.~\ref{fig:hardware_upgrade}(d)) have been replaced with new embedded electronics manufactured by Apptronik. These new control boards are equipped with, a powerful micro-controller, a TI Delfino, that performs complex computations with low-latency for signal processing and control. The control boards are installed in a special board case (Fig.~\ref{fig:hardware_upgrade}(b)) holding safely all cables connected to the board. This wiring routing and housing detail is important because Mercury hits the ground hard when walking in rough terrains and performs experiments by being hit by people and balls. It secures signal and power cables to enable solid signal communications. 

Thirdly, we manufactured carbon shells (Fig.~\ref{fig:hardware_upgrade}(c)) to reinforce the thigh linkages. We also redesigned the robot's shank to increase structural stiffness by including two carbon fiber cylinders as supporting linkages. In addition, we designed new passive feet in the form of thin and short prisms that are a few centimeters long. The feet pivot about a pin fulcrum which connects in parallel to a spring between the foot support and the pivoting ankle. A contact switch is located on the front of the foot and engages when the foot makes contact with the ground (see Fig.~\ref{fig:hardware_upgrade}(e) for mechanical details). These contact switches are used to terminate swing foot motion controls when the swing foot touches the ground earlier than anticipated. The main purpose of the line feet is to prevent yaw rotations of the entire robot turning around the supporting foot. Previously, our robot had quasi-pointed feet, which caused the robot's heading to turn due to any vertical moments. The mechanical line contacts provided by the passive feet interact with the ground contacts as a friction moment preventing excessive body rotations. 

\subsection{Challenges in Passive-Ankle Locomotion}
\begin{figure}
\centering
\includegraphics[width=\columnwidth]{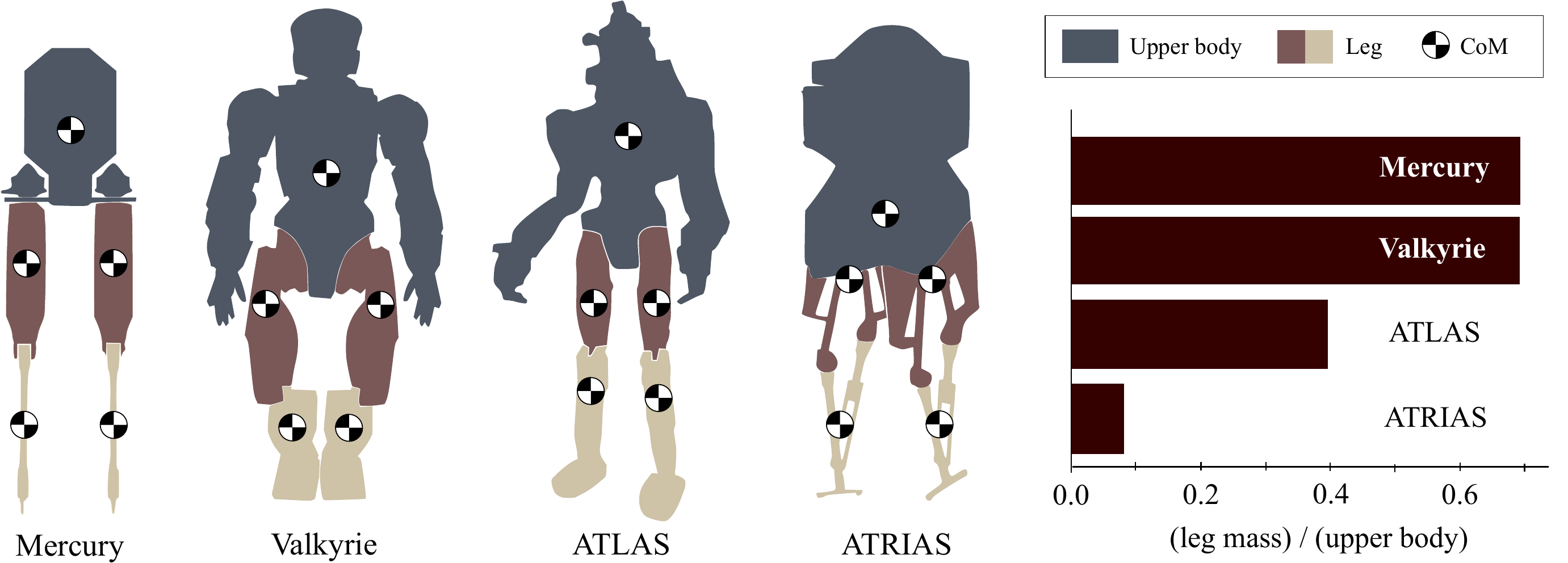}
\caption{{\bf Mass distribution of various biped robots.} Notice that the the robots shown here are not scaled equally. They are shown to compare the relative location of their CoM. The CoM locations are superimposed on the upper-body and leg links of each robot. The bar graph depicts the ratio of the total leg mass over the upper body mass. Notice that ATRIAS has a mass distribution different than typical humanoids by having the torso CoM near the hip joint and a small leg-to-upper-body mass ratio.}
\label{fig:mass_distribution}
\end{figure}

To discuss the locomotion challenges presented by Mercury, it is necessary to discuss the mass distribution of Mercury against other bipeds (Fig.~\ref{fig:mass_distribution}). The robots' inertia information used for this comparison is taken from open source robot models found in the following public repositories: \url{https://github.com/openhumanoids} (Valkyrie), \url{https://github.com/dartsim/} (ATLAS), and \url{https://github.com/sir-avinash/atrias-matlab} (ATRIAS). Mercury's mass distribution is somewhat similar to anthropomorphic humanoid robots such as Valkyrie \cite{Radford:ca} or Atlas \cite{Kuindersma:2015cw}. These robots have (1) a torso CoM located around the center of its body, and (2) the ratio between the total leg mass and the torso mass is significant, about 0.4. On the other hand, ATRIAS \cite{8360164} has a mass distribution optimized to be a mechanical realization of the inverted pendulum model, which is designed to aid with the implementation of locomotion controllers. Unlike other humanoid robots, ATRIAS's torso CoM location is close to the hip joints and the ratio between the total leg mass to the torso is negligible, which is less than 0.1. 

While Mercury and ATRIAS are similar in their lack of ankle actuation and number of DoFs, the difference in mass distributions creates difficulties on locomotion control. Since ATRIAS has its torso CoM close to the hip joint axis, the link inertia reflected to the hip joint is small, which reduces the difficulty of controlling the robot body's orientation. In contrast, the CoM of Mercury and other humanoid robots mentioned above are located well above the hip joint, which creates a larger moment arm and increases the difficulty of body orientation control. 

Next, since ATRIAS has negligible leg mass compared to its body, body perturbations caused by the swing leg are also negligible. However, Mercury, having significant leg mass, causes noticeable body perturbations during the swing phase. Thus, it becomes necessary for Mercury to have a whole-body controller which can compensate against Coriolis and gravitational forces introduced by the swing leg to maintain desired body configurations, follow inverted pendulum dynamics, and control the swing foot to desired landing locations. Overall, in addition to Mercury's SEAs and lack of ankle actuation, its mass distribution makes it more difficult to control.

\section{Implementation Details}
\label{sec:implementation}


\subsection{Walking Control}
\label{sec:walking_motion}

\begin{figure}
    \centering
    \includegraphics[width=0.9\columnwidth]{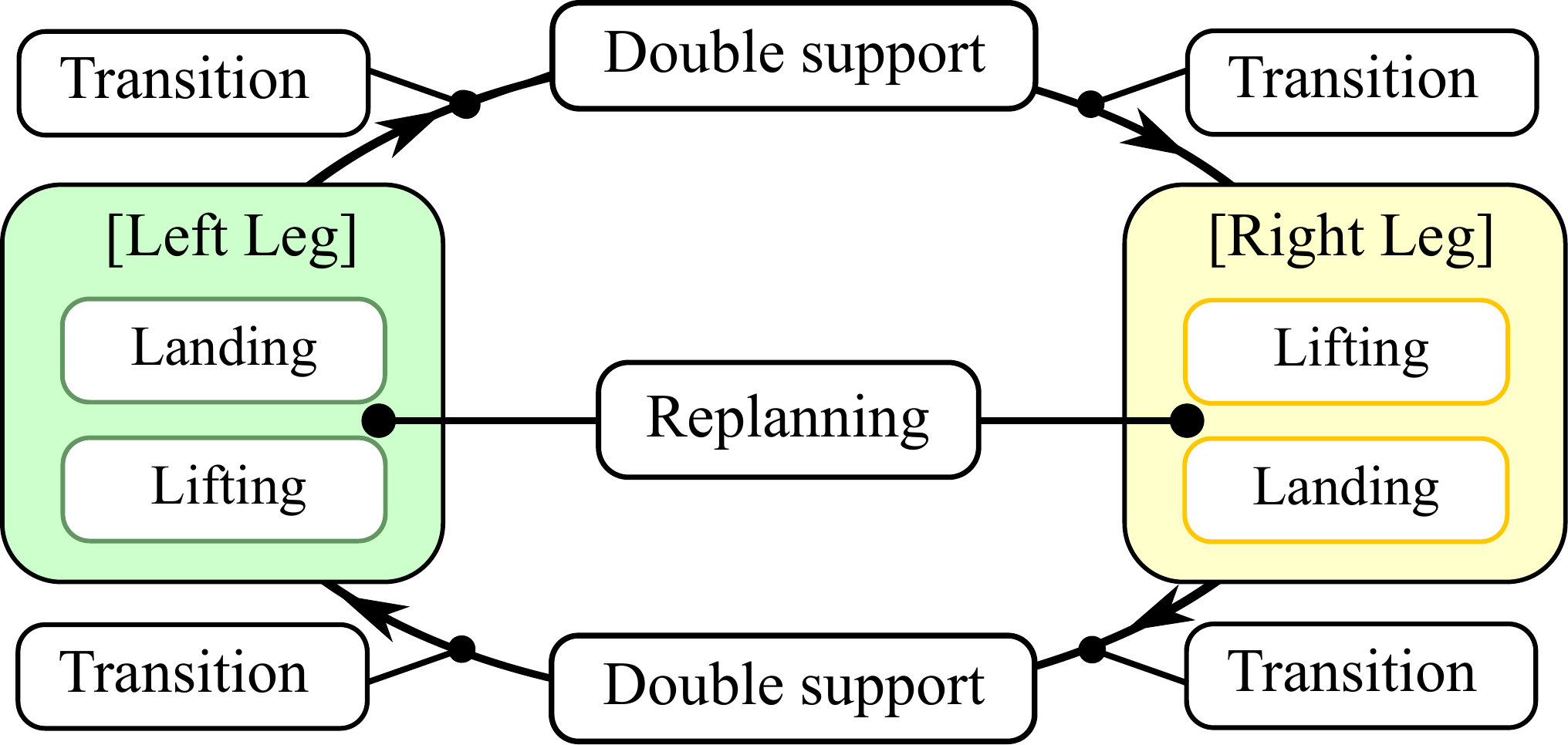}
    \caption{{\bf State Machine.} Our biped walking motion is achieved via sequential contact phase changes governed by the temporal parameters shown in Table.~\ref{tb:walking_time}. The robot's swing leg phase can be terminated earlier than the predefined swing time if the contact is detected before the end of the swing phase. In the middle of the swing, the next foot placement is computed by the TVR planner.}
    \label{fig:state_machine}
\end{figure}

For our purposes, a biped's walking control process consists of three phases: swing (or single stance), double stance, and contact transition. In particular, the contact transition ensures smooth transition from single to double contact. Each phase starts and ends following predefined temporal parameters as shown in Table.~\ref{tb:walking_time}. The swing phase can, and it often does, terminate earlier than the specified swing time because the biped might make contact with the ground earlier than planned. We automatically terminate the swing phase upon detecting contact to prevent sudden jerks that can occur when pushing against the ground. The ground contact is detected by the limit switches attached to the spring-loaded passive ankles shown in Fig.~\ref{fig:hardware_upgrade}(e). The locomotion phases are illustrated in Fig.~\ref{fig:state_machine}. At the middle of the duration of each swing phase, our TVR planner computes the immediate foot step location to achieve stable locomotion based on the policy given by Eq.~\eqref{eq:tvr_step_eq}. This decision process works as follows. After breaking contact with the ground, the swing foot first moves to a predefined default location with respect to the stance foot. Then, a new foot landing location is computed using the TVR planner. Based on this computation, the swing trajectory is re-adjusted to move to the computed foot landing location completing the second half of the swing motion until contact occurs.

Due to the non-negligible body-to-leg-weight ratio, when the swing motion occurs, it disturbs the robot's body. As the inertial coupling between the leg and body has a strong negative effect on the robot's ability to walk and balance, it is important to reduce these types of disturbances. In particular, we mentioned earlier that the robot's swing leg first moves to a default location, and from there it computes a new foot landing location to dynamically balance and walk. Therefore, we focus on reducing the jerky motion that occurs from re-adjusting the foot trajectory at the middle of the swing motion. In our experiments, we first move to the default swing location using a B-spline, and then compute a minimum jerk trajectory to achieve the final landing location. The inclusion of this minimum-jerk trajectory is important as it significantly reduces the said disturbances between the swinging leg and the robot's body posture.

When the swing motion ends, the state machine switches to the contact transition phase. Here the DynWBC control block shown in Section.~\ref{sec:DynWBC} plays a key role to smoothly transition the contact from single to double support without introducing additional jerky movement. On the other hand, when a contact occurs, triggering a switch from single to double support, the KinWBC control block can generate a discontinuity of the joint position command. To reduce this additional jerk caused by KinWBC, the joint position command of the swing leg at the end of the swing phase is linearly interpolated with the command from KinWBC for the transition phase. As the contact transition progresses, the ratio between the final joint position command and the transition phase decreases which completes the transition. By doing all of these improvements, we accomplish smooth motions with reduced jerk for effective walking. 
\begin{table}
\centering
\begin{tabular}{>{\centering}m{0.27\columnwidth} %
                >{\centering}m{0.27\columnwidth} %
                >{\centering}m{0.27\columnwidth} %
                @{}m{0pt}@{}}
\specialrule{1.5pt}{1pt}{1pt}
{Double stance}
& {Transition}
& {Swing}
&\\[1mm] 
\hline
\hline 
0.01 sec & 0.03 sec & 0.33 sec 
&
\\[1.5mm]
\hline
\end{tabular}
\caption{Temporal parameter of walking}
\label{tb:walking_time}
\vspace{-1.5mm}
\end{table}

\begin{table}
\centering
\begin{tabular}{>{\centering}m{0.27\columnwidth} %
                >{\centering}m{0.27\columnwidth} %
                >{\centering}m{0.27\columnwidth} %
                @{}m{0pt}@{}}
\specialrule{1.5pt}{1pt}{1pt}
{Double stance}
& {Transition}
& {Swing}
&\\[1mm] 
\hline
\hline 
$R_x$, $R_y$, $z$ & $R_x$, $R_y$, $z$ & $R_x$, $R_y$, $z$ 
& \\
\hline 
- & - & ${\rm Foot}_{x, y, z}$ 
& 
\\[1.5mm]
\hline
\end{tabular}
\caption{Task Setup}
\label{tb:task_setup}
\vspace{-1.5mm}
\end{table}

\begingroup 
\setlength\arraycolsep{2pt}
\begin{table*}
\centering
\begin{tabular}{>{\centering}m{0.05\columnwidth} %
                >{\centering}m{0.5\columnwidth} %
                >{\centering}m{0.75\columnwidth} %
                >{\centering}m{0.5\columnwidth} %
                @{}m{0pt}@{}}
\specialrule{1.5pt}{1pt}{1pt}
&{Double support}
& {Transition (right)}
& {Swing (right)}
&\\[1.2mm] 
\hline
\hline 
$\bm{W}_{\ddot{\mathbf{q}}}$&  $10^2  \times \mathbf{1}_{12\times1}$ & $10^2 \times \mathbf{1}_{12\times1}$ & $10^2 \times \mathbf{1}_{12\times1}$ 
& \\[2.mm]
\hline 
$\bm{W}_{r}$& 
$\begin{bmatrix}1, & 1,& 0.01,& 1,& 1,& 0.01\end{bmatrix}^{\top}$  & $\begin{bmatrix}1\rightarrow 5, & 1 \rightarrow 5, & 
0.01 \rightarrow 0.5,& 1,& 1,& 0.01\end{bmatrix}^{\top}$ & 
$\begin{bmatrix}5, & 5,& 0.5,& 1,& 1,& 0.01\end{bmatrix}^{\top}$
& \\[2.mm]
\hline 
$\bm{W}_{c}$& 
$10^3\times \mathbf{1}_{6\times1}$ & 
$\begin{bmatrix}\left(10^{3} \rightarrow 10^{-3}\right)\times\mathbf{1}_{1\times3}, & 10^{3}\times\mathbf{1}_{1\times3}\end{bmatrix}^{\top}$ & 
$\begin{bmatrix} 10^{-3}\times\mathbf{1}_{1\times3}, & 10^{3}\times\mathbf{1}_{1\times3}\end{bmatrix}^{\top}$ 
&
\\[2.mm]
\hline
\end{tabular}
\caption{{\bf Weight Setup.} Here, the values of the weight matrices are described in vector form because we consider only diagonal weight matrices. The components associated with reaction and contact weights are six dimensional, starting from the right foot's $x$, $y$, and $z$ directions and then considering the left foot's Cartesian components; therefore, $\bm{W}_r$ and $\bm{W}_c$ have six components.}
\label{tb:weight}
\vspace{-1.5mm}
\end{table*}
\endgroup 

\subsection{Task and Weight Setup of WBLC}
\label{sec:WBLC_setting}
The WBLC task setup for each phase is summarized in Table.~\ref{tb:task_setup}. A common task for every control phase is the body posture task which keeps the body's height, roll, and pitch constant. Since Mercury has only six actuators, the robot cannot directly control its body yaw rotation and horizontal movement most of the time. Therefore, we only control three components ($R_x$, $R_y$, $z$) of the six-dimensional body motions ($R_x$, $R_y$, $R_z$, $x$, $y$, $z$). Here $R_{\{\cdot\}}$ stands for rotations.

During the swing phase, we control the linear motion of the foot in addition to the robot's body posture. The swing foot task is hierarchically ordered under the body posture task to prevent the swing motion from influencing the body posture control. However, this priority setup is not enough to completely isolate the body posture control from the swing motion control because the null space of the body task does not remove the entire six DoF body motion. In our case, the body posture control task only controls three of the six-dimensions of body motion, which means that the other three components still reflect on the swing foot task even after the foot task has been projected into the null-space of the body posture task. To further decrease the coupling between the body motion and the swing foot intended motion, we set to zero all of the terms corresponding to the floating base DOFs appearing in the foot task Jacobian. By doing so, the three actuators in the stance leg are dedicated only to body posture control while the other three actuators in the swing leg are dedicated to control the swing foot trajectory. 

The values of the weights of the cost function in Eq.~\eqref{eq:qp_cost_wblc} in DynWBC are specified in Table.~\ref{tb:weight}. These values are presented in vector form because all of the cost matrices are diagonal. $\bm{W}_{\ddot{\mathbf{q} } }$ is the weight matrix for relaxing desired joint accelerations to adjust for partially feasible acceleration commands. These weights are set to relatively large values to penalize deviations from the commanded joint accelerations. The same values are kept for every phase. 

$\bm{W}_{r}$ and $\bm{W}_{c}$ change as a function of the walking control phase because the reaction forces and feet movements are regulated by those weights. During the double support phase, the weights related to the contact point acceleration, $\bm{W}_{c}$, are assigned a large value, $10^3$. Penalizing contact accelerations approximates contact conditions without imposing hard constraints. 
Also during double support, the weight matrix regulating reaction forces, $\bm{W}_{r}$, is assigned relatively small values to provide sufficiently large forces to support the robot's body. Note that  $\bm{W}_{r}$ penalizes the tangential direction values more than the normal direction values, which helps to fulfill the friction limits associated with the contact reaction forces. 

$\bm{W}_r$ and $\bm{W}_c$ change value during the contact transition phase. The right arrows in Table \ref{tb:weight} indicates that the weights transition smoothly from the left to the right values. For instance, $1\rightarrow 5$ means that the value applied to the weight is set to $1$ at the beginning of the transition phase, and we linearly increase it to $5$ by the end of the phase. Let's take the example with the right foot during the transition phase. At the beginning of the transition phase the weight values coincide with the values in the previous phase, i.e. double support. At the end of the transition phase, when the right leg is about to leave the ground and start the swing phase, the first three terms of $\bm{W}_r$, coinciding with the Cartesian components of the right foot reaction force, are set to large values to penalize reaction forces. At the same time, the three first terms of $\bm{W}_c$ are set to tiny values to boost swing accelerations on the right foot. 

During this transition we perform an additional step. For the constraint defined in Eq.~\eqref{eq:wblc_frz_bound} of DynWBC, i.e. $\bm{S}\mathbf{F}_{r} \leq \mathbf{F}_{r,z}^{\rm max}$, we linearly decrease the value of the upper bound $\mathbf{F}_{r,z}^{\rm max}$ to drive the right foot normal force to zero before the swing motion initiates. This linear decrease starts with the value set during double support and ends with a value equal to zero.

\subsection{Base State Estimation}
\begin{figure}
    \centering
    \includegraphics[width=\columnwidth]{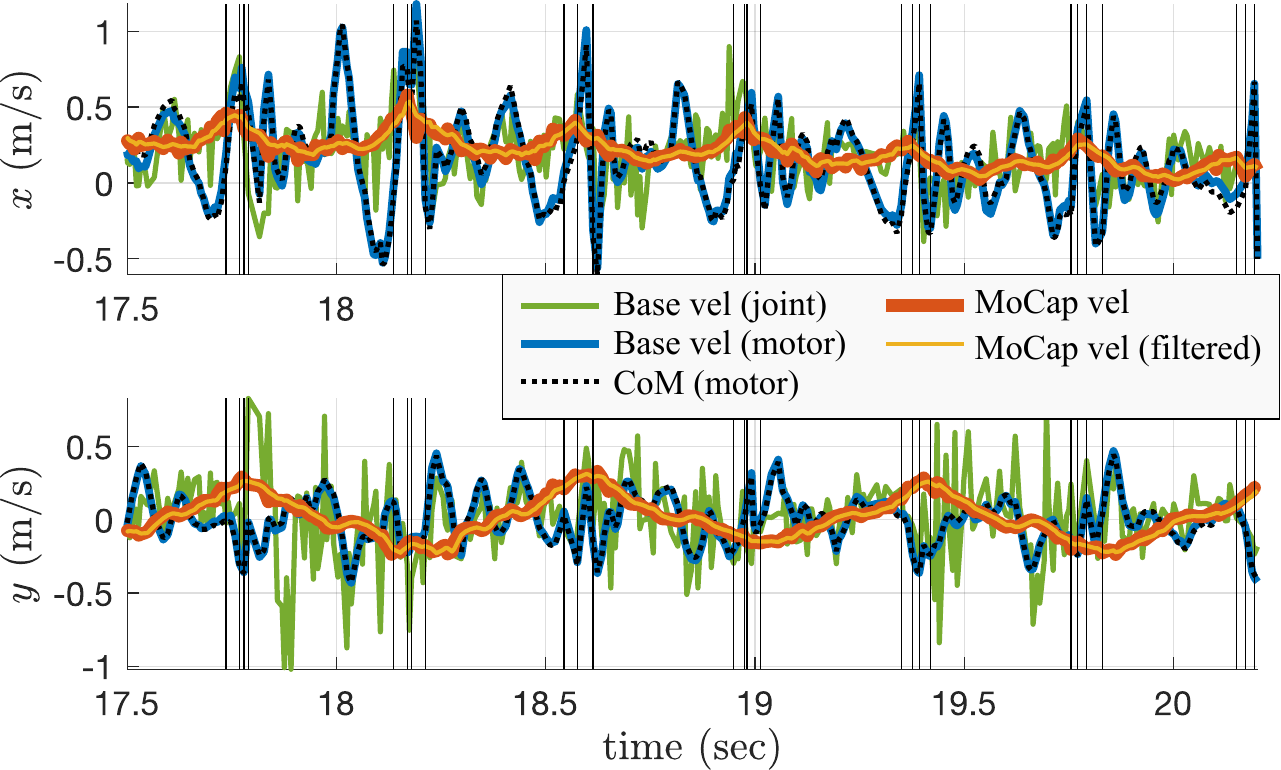}
    \caption{{\bf Base and CoM velocity.} The base and CoM velocity from different measurements are plotted. The base velocities are computed with joint velocity data measured by absolute encoders or quadrature encoders. The base velocity estimated by absolute encoders are too noisy and significantly fluctuates during the swing phases. Even with quadrature encoders, the fluctuation remains although the noisy level is lower than the ones estimated by using absolute encoders. During experiments, we use the results indicated by the yellow line, which corresponds to the filtered velocity data obtained from the MoCap system.}
    \label{fig:base_vel}
\end{figure}

As the true CoM state is subject to errors from the model and disturbances from the swing leg motion, our current implementation instead uses the robot's base state, and assumes that the CoM of the robot is approximately at this location. The robot's base is a concrete point on the torso indicated by a black dot in Fig.~\ref{fig:configuration}. The base point was chosen by empirically comparing the CoM position and base position to find the lowest discrepancies. Fig.~\ref{fig:base_vel} shows velocity estimation values. As we can observe, the difference between the CoM velocity (black dotted line) and base velocity (blue solid line) is unperceivable. This enables us to (1) decouple the computation of the CoM state from the swing leg motion, and (2) perform a straight-forward sensor-fusion process with a Kalman filter by combining the sensed body positions from joint-encoders and the overhead MoCap system. 

As said, Fig.~\ref{fig:base_vel} compares the base velocity data obtained from different sensors. In Section.~\ref{sec:robot_description}, we stated that there are two ways to measure joint data: one is using the absolute encoders directly attached to the robot joints, and another one is using the quadrature encoders attached to the back of the electric motors by multiplying their value with the actuator's transmission ratio. The green lines and the blue lines on the above figures correspond to the base velocities computed from data measured by absolute encoders and quadrature encoders. The blue lines are less noisy, but both green and blue data are not proper for our walking planner because the velocity profile shows a significant fluctuation, which makes the prediction of the state challenging. However, the velocity data obtained from the MoCap system, i.e. the red lines, shows a consistent trend with the walking phases such that we decided to rely on it. To deal with MoCap marker occlusions, we perform sensor fusion between the MoCap and encoder data via Kalman filtering and average filtering techniques. This data is shown as a yellow line on the previous figure showing that it is fairly similar to the red line. 

For the estimation of the base positions in global frame we use the MoCap system. As for estimating base positions with respect to the stance foot we rely only on the robot's IMU and joint encoder data without using the MoCap system. This last process is more robust than attaching LED sensors to the feet because they incur frequent occlusions and break often due to the repetitive impacts.   

\subsection{Kinematic Model Verification With MoCap Data}
As we mentioned in the previous section, an accurate kinematic model is very important to compute stabilizing foot landing locations via the TVR planner. Moreover, for real-time WBLC, the model's accuracy significantly influences the landing location accuracy. To perfect our kinematic model which was initially built using the parameters obtained from CAD design, we utilize the MoCap system. By comparing the MoCap data and the kinematic model data, we tune the model parameters and enhance the accuracy of the kinematic model until the two sets of data are sufficiently close together. 
\begin{figure}
    \centering
    \includegraphics[width=\columnwidth]{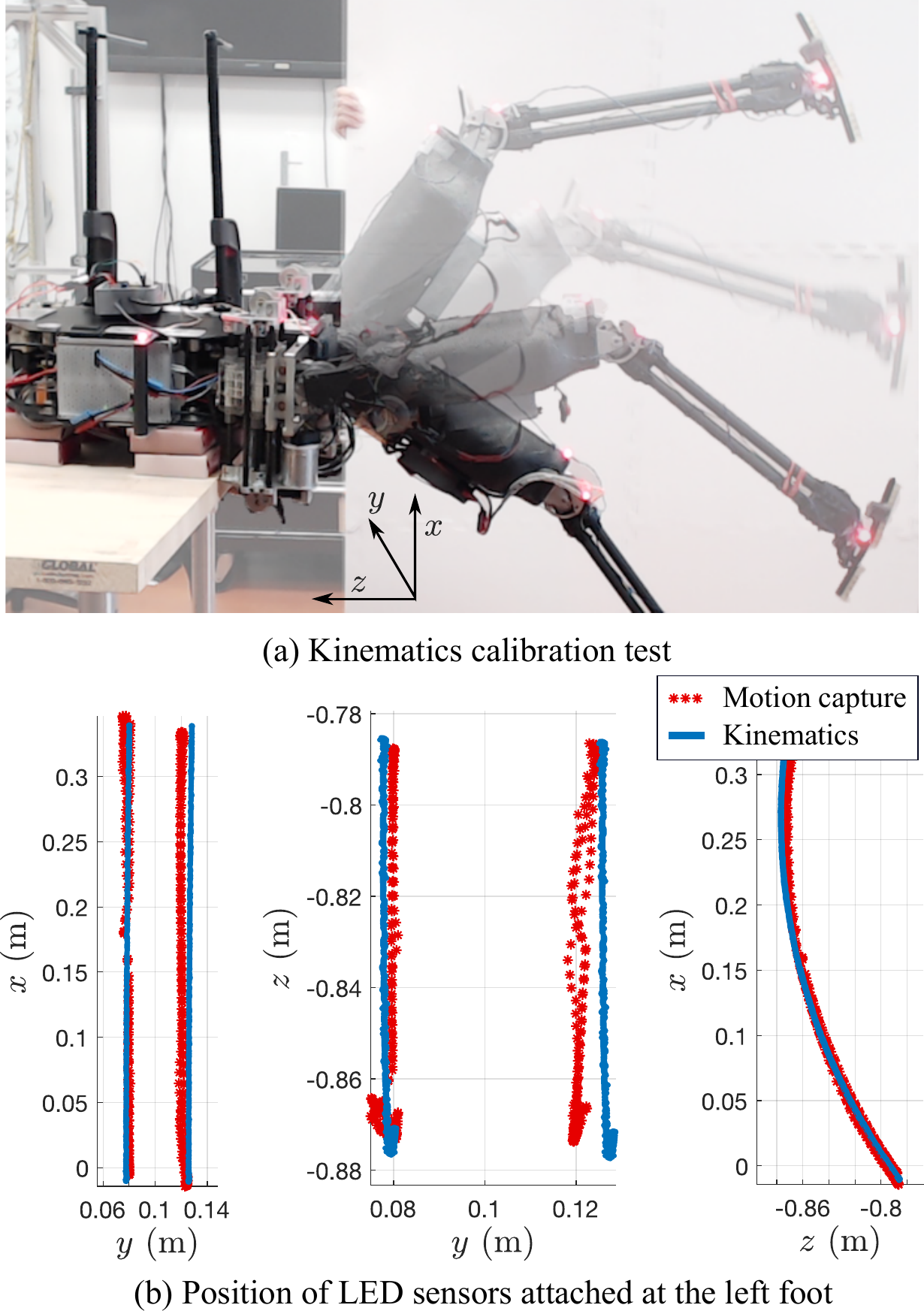}
    \caption{{\bf Kinematic model calibration.} (a) Mercury swings its leg while its torso is fixed on a table. (b) To tune the kinematic model parameters, we compare the LED position data obtained from the MoCap system and the position data computed by the kinematic model.}
    \label{fig:kin_calib}
\end{figure}

For this calibration process, we first fix Mercury's torso on top of a table as shown in Fig.~\ref{fig:kin_calib}(a). For this fixed posture, we let the robot swing one of its legs and simultaneously gather MoCap and kinematic data. The positions of the LED sensors attached to the leg are post-processed to be described in the robot's local frame, which is defined by three LED sensors attached to the robot's body (see Fig.~\ref{fig:configuration}). The two different sets of data, one obtained from the MoCap system and the other one obtained from the current robot kinematic model are used to further tune the kinematic parameters. Fig.~\ref{fig:kin_calib}(b) shows both the LED position data measured by the MoCap system and the same position data measured by the joint encoders using the tuned kinematic model. The result shows that the error of our final kinematic model has less than a 5\unit{mm} error. 

\section{Results}
\label{sec:experiment}
We conducted extensive walking and stepping experiments of various kinds using our passive ankle biped robot, Mercury. For all of these experiments, Mercury was unsupported, that is, without overhead support. The experiments show stable behavior during directional walking, push recovery, and mildly irregular terrain walking. We also deploy the same control and dynamic walking schemes to our new lower body humanoid robot, DRACO, and rapidly accomplished dynamic balancing. Finally, we conducted simulations using other humanoid robots to show the versatility of our whole-body controller and walking algorithm.

\begin{figure*}
\includegraphics[width=2\columnwidth]{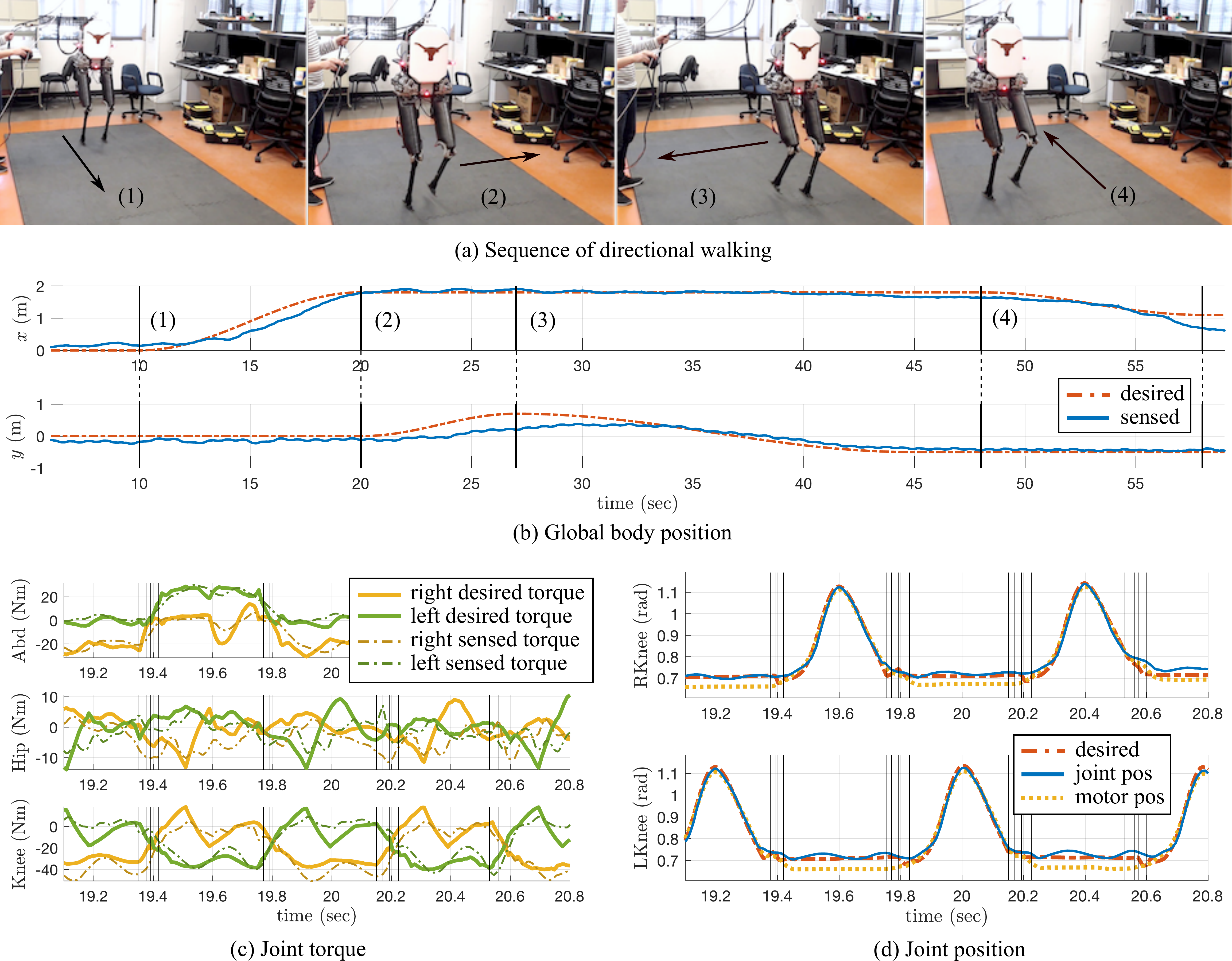}
\caption{{\bf Directional walking of Mercury.} (a) Mercury starts walking from the back and goes forward, left, right, and back again. (b) The robot's location from the base LED sensor is compared with the commanded walking trajectory. Mercury follows fairly well the desired trajectory but shows limited convergence rate with respect to moving towards the lateral direction. (c) Joint torques change smoothly despite quick contact transitions thanks to our WBLC method. (d) Knee joint position data shows that spring deflection models are effective for reducing joint position tracking error.}
\label{fig:directional_walking}
\end{figure*}

\subsection{Directional Walking}
Directional walking means achieving dynamic walking toward a particular direction. To achieve this, we manipulate the origin of Mercury's reference frame. In turn, our TVR planner controls Mercury's foot stepping to converge to the reference frame, which for this test is a moving target. In other words, we steer the robot in the four cardinal directions in this manner, see Fig.~\ref{fig:directional_walking} (a). Fig.~\ref{fig:directional_walking} (b) shows the time trajectory of the desired robot's path and the actual robot's location. The actual location is obtained using the MoCap system based on the LED attached to the robot's base. These results show that Mercury follows the commanded path relatively well albeit slow convergence rates in the lateral direction possibly due to the limited hip's abduction/adduction range. 

Fig.~\ref{fig:directional_walking} (c) shows commanded and sensed joint torque data. The vertical black lines indicate the walking control phases. As we can see, the torque commands smoothly transition despite contact changes. The knee torque commands change between 0 and 40 \unit{Nm} depending on the control phase of the leg, but there is no discontinuity causing jerky behavior of the desired torque commands despite the short (0.06 \unit{sec}) transition periods. 

The right and left knee joint position data are shown in Fig.~\ref{fig:directional_walking} (d). As mentioned in Section.~\ref{sec:joint_ctrl}, the desired motor position commands are adjusted to account for spring deflections. The data shows that joint positions sensed with the absolute encoders are close to the position commands while the motor position data is off by the amount corresponding to spring deflections. The spring deflection compensation is notable when the knee joint supports the body weight, i.e. the periods between $19.8 \sim 20.2\unit{sec}$ for the right knee and $19.4 \sim 19.8\unit{sec}$ for the left knee.

\begin{figure*}
    \centering
    \includegraphics[width=2\columnwidth]{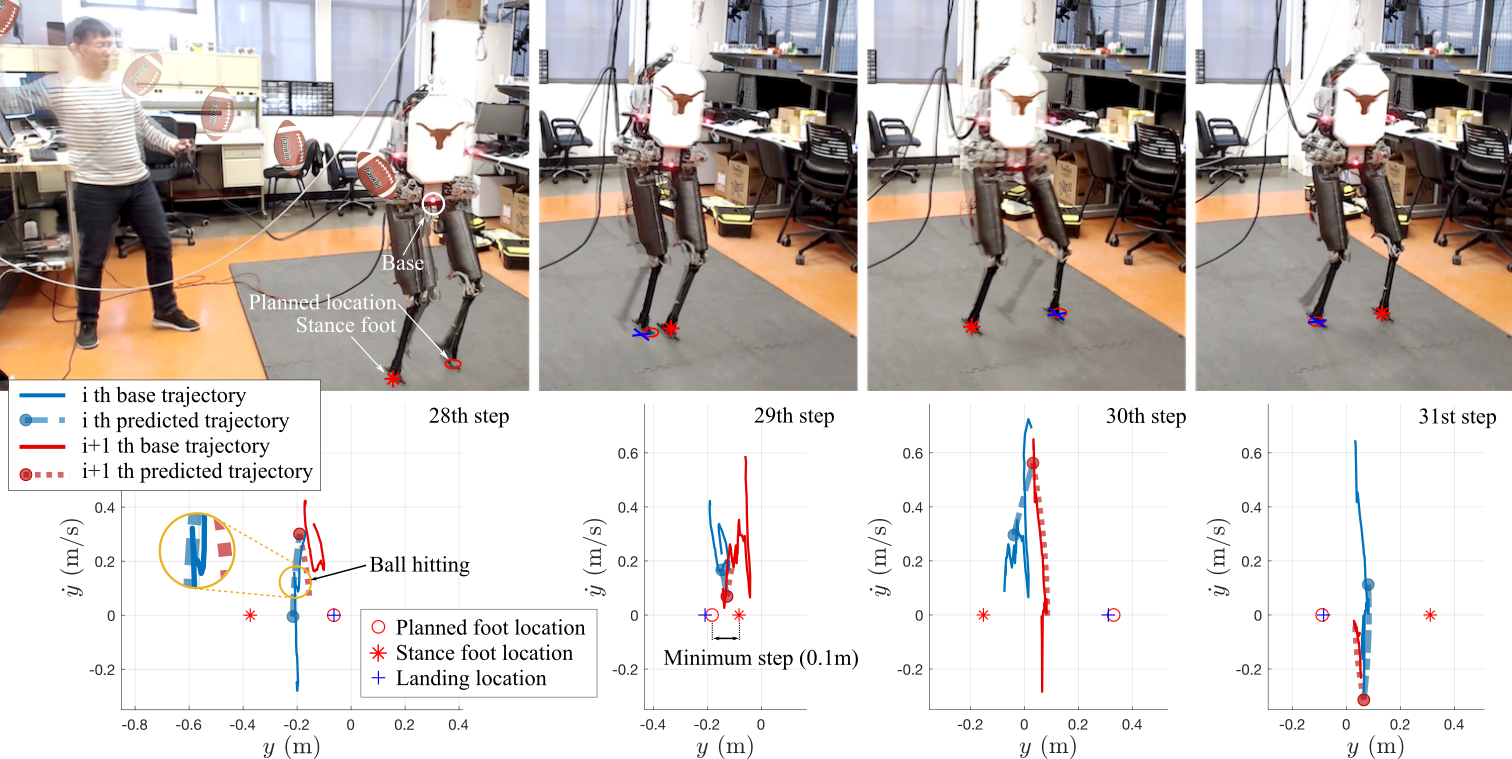}
    \caption{Mercury recovers its balance after being disturbed by a lateral impact applied by throwing a junior American rubber football ball, weighting $0.32\unit{kg}$. In the 28th step, the ball hits Mercury on its side as depicted in the lateral change of the CoM state, i.e. $y$ direction. For this instance, when the lateral impact happens, the next foot landing location, in our case the left leg, has already been planned and there is nothing else that can be done. So Mercury finishes the lateral step without responding to the disturbance. For the following step, step 29th, the CoM velocity is positive in the $y$ direction due to the lateral disturbance. This value on the CoM state causes our TVR walking planner to trigger a recovery step using the right foot which is commanded to move inward towards the stance foot. However, the amount it has to move would cause a collision with the stance leg, therefore our planner chooses to land the right foot at the minimum lateral range of 10\unit{cm} from the stance leg. This choice, causes the robot to only partially recover from the disturbance but failing to reverse velocity. As a result, for the next step, step 30th, Mercury's TVR walking controller decides to take a large step, 48\unit{cm} from the stance leg, which enables it to reverse velocity in the direction opposite to the impact. Finally, Mercury goes back to its nominal balancing motion in step 31st.}
    \label{fig:throwing_ball}
\end{figure*}

\subsection{Robustness Of Balance Controller}

\begin{figure}
    \centering
    \includegraphics[width=\columnwidth]{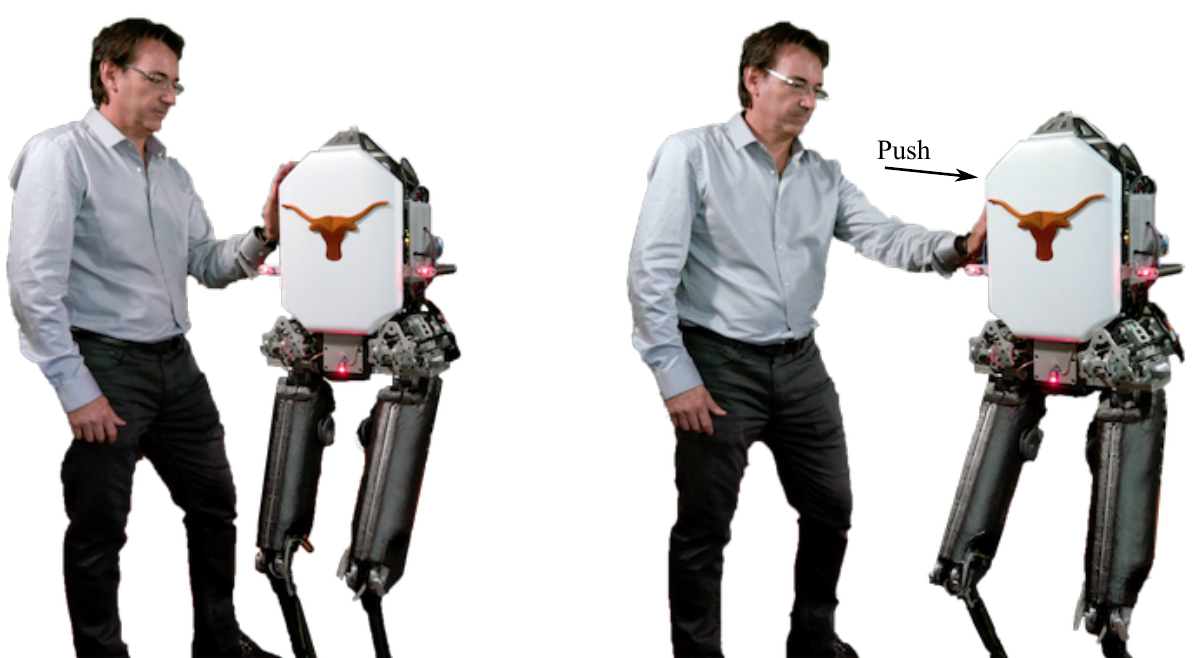}
    \caption{{\bf Interaction with a human subject.} Mercury maintians its balance despite the continuous pushing forces.}
    \label{fig:human_push}
\end{figure}

\begin{figure*}
    \centering
    \includegraphics[width=2\columnwidth]{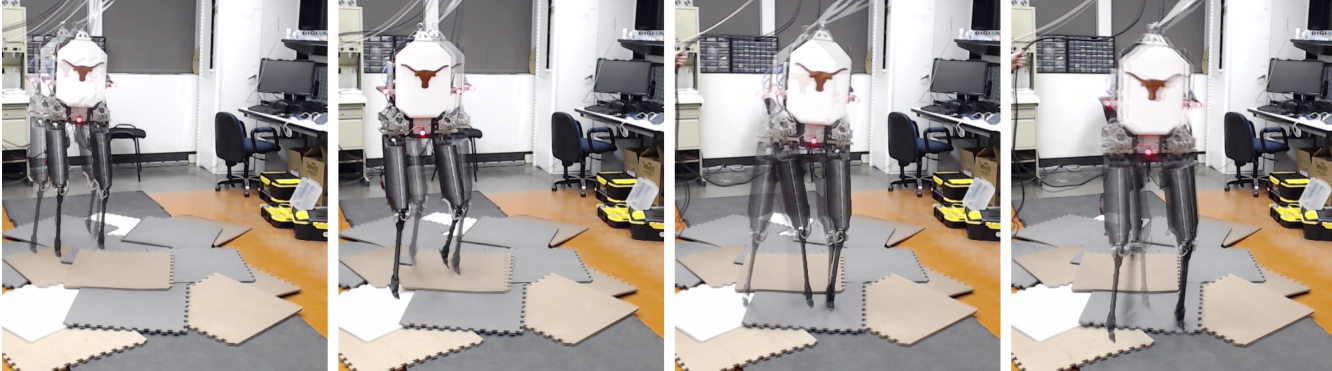}
    \caption{{\bf Forward walking over an irregular terrain.} Mercury walks forward over an irregular terrain constructed with foam mats arranged on top of each other. The robot's feet sometimes slip over the mat segments since the latter do not stick tightly to each other. Therefore there are multiple disturbances. Our control and walking algorithms accomplish the necessary robustness to traverse these type of terrain including height variations of 2.5 \unit{cm}), foot slippage, and foot trippings.}
    \label{fig:rough_terrain}
\end{figure*}

To demonstrate the robustness of the proposed walking control scheme, we conducted multiple instances of an experiment involving external disturbances. The first test, shown in Fig.~\ref{fig:throwing_ball}, analyzes Mercury recovering its balance after a junior football ball of weight $0.32\unit{kg}$ and horizontal speed of about $9 \unit{m/s}$ impacts its body. A second test, shown in Fig.~\ref{fig:human_push}, shows a person continuously pushing Mercury's body with gentle forces to see how the robot reacts. Finally, the last experiment, shown in Fig.~\ref{fig:rough_terrain}, shows Mercury walking in a mildly irregular terrain without knowledge or sensing of the terrains topology. In the three experiments, Mercury successfully recovers from the disturbances.

For the ball impact experiment shown in Fig.~\ref{fig:throwing_ball}, we show the phase plots of the lateral CoM direction. Since the ball hits the robot laterally, the analysis is done on the $y$ direction. Lateral impact recovery is difficult because the hip abduction/adduction joints have a very limited range of motion, $\pm 15 \unit{^o}$. Due to the very small width of the feet, the landing location has to be very accurate as previously discussed. Each phase plot in this figure, shows two sequential steps, depicted in blue and red color lines. For instance, for the 28th step, we differentiate the solid blue line, which represents the sensed base trajectory for the actual 28th step, from the solid red line, which representss the trajectory for the next step, the 29th. Dotted blue and red lines represent the predicted trajectory given the TVR control policy and pendulum dynamics hypothesis. The particular operating details of the TVR controller during this impact experiment are described in the caption of Fig.~\ref{fig:throwing_ball}. In essence, the ball hits the robot at the 28th step and at the 30th step, Mercury fully recovers its balance, going back to the normal regime at the 31st step. 

Also from Fig.~\ref{fig:throwing_ball}, we analyze the foot landing accuracy. In the phase plots, the red star, the red circle, and the blue cross represent the stance foot, commanded foot landing location, and actual foot landing location, respectively. Except during the recovery steps, 29th and 30th steps, the foot landing location errors are less than 0.5 \unit{cm}, as seen in the 28th and 31st steps. This is significantly less error than the maximum tolerable one as shown in the uncertainty analysis of Fig~\ref{fig:phase_plot_analysis}. In analyzing extended experimental data, the foot landing error is consistently less than 0.5 \unit{cm}.

Our control and walking methods are robust to mild terrain variations as shown in Fig.~\ref{fig:rough_terrain}. For this particular experiment, we set $\kappa_x$, shown in Table.~\ref{tb:planner_param}, to a value of $0.25$ to enable the robot to keep moving forward despite the terrain variations. In addition, the robot's feet sometimes get stuck on the edge of the mats, which adds difficulty to the locomotion process. However, the robot successfully traverses the terrain. 

\subsection{Experimental Evaluation On New Biped Robot DRACO}
\begin{figure*}
    \centering
    \includegraphics[width=2\columnwidth]{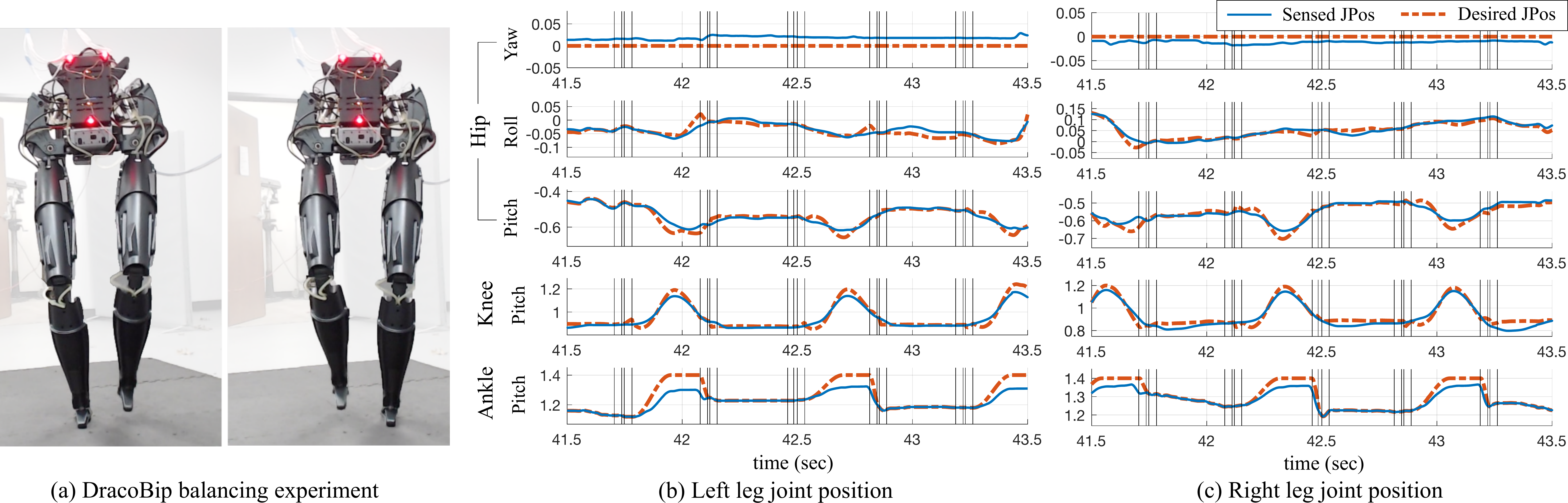}
    \caption{{\bf Balancing experiment of DRACO.} Using the same WBLC and TVR algorithms from Mercury, DRACO was able to balance unsupported within a few days.}
    \label{fig:draco_exp}
\end{figure*}

DRACO is our newest humanoid lower body, having ten viscoelastic liquid-cooled actuators \cite{Kim:2018ep} on its hips and legs. Each limb has five actuators: hip yaw, roll, pitch, knee pitch, and ankle pitch. The IMU is the same as in Mercury, a STIM 300, and the MoCap LED sensor system is configured similarly to Mercury. The robot has many interesting features such as liquid cooling, tiny feet, quasi-passive ankles, and elastomers in the actuator drivetrains. We won't describe the hardware details of DRACO as they are being prepared for submission for an upcoming paper.

To equate DRACO to Mercury in some respects, we apply a soft joint stiffness policy to the ankle pitch emulating a passive joint. For this first experiment, we set the hip yaw joint to a fix position with a joint control task implemented in WBLC. From a controller's standpoint, Mercury and DRACO are very similar for this experiment. DRACO is forced to perform dynamic locomotion without controlling ankles in similar ways than Mercury. For now, we check for foot contacts on DRACO based on ankle joint velocity measurements. As shown in Fig.~\ref{fig:draco_exp}, DRACO balances successfully unsupported, just like Mercury did.

For WBLC on DRACO, we generated the robot's model using the CAD files and slightly adjusted mass values from gravity compensation tests. Except for feedback gains of the joint position controllers, we use similar planner parameters to Mercury. $t'$ is set to $[0.21, \: 0.2]$ and $\kappa$ is set to $[0.08, \: 0.13]$ for the experiment. Testing on DRACO was successfully accomplished thus demonstrating that our WBLC-TVR framework is easily transferable to multiple robots, showing the generality of our methods.

\subsection{Simulation Results in Assorted Platforms}
\begin{figure}
\centering
 \includegraphics[width=\columnwidth]{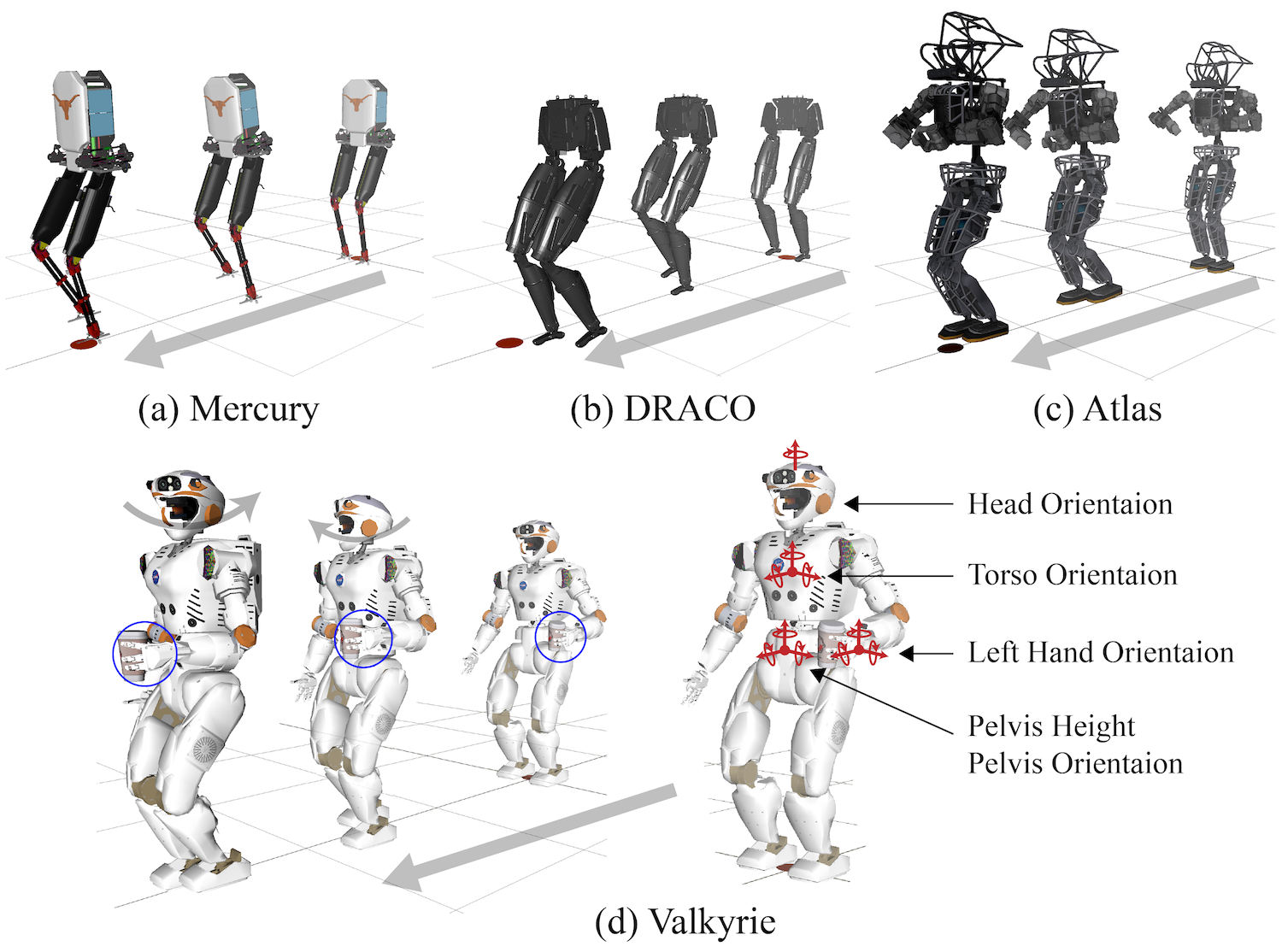}
\caption{{\bf Simulations using four robotic platforms: Mercury, DRACO, Atlas, and Valkyrie.} The foot task is present in all of the simulations. DRACO contains an additional hip position task. Atlas and Valkyrie have additional tasks such as pelvis and torso orientation, foot orientation, and arm joint configuration. Lastly, Valkyrie has an additional head orientation and left hand orientation tasks. (a), (b), and (c) show the simulation results for dynamic walking using Mercury, DRACO, and Atlas. (d) shows locomanipulation capabilities of our WBLC performed by Valkyrie.}
\label{fig:sim_multi_robots}
\end{figure}

To show further applicability of the proposed control methods, we implement and test our WBLC and TVR algorithms on assorted robotic platforms such as Mercury, DRACO, Atlas, and Valkyrie. We implemented two types of simulation scenarios: dynamic walking and locomanipulation. Mercury, DRACO, and Atlas are utilized to implement dynamic walking motions. As mentioned in Section $\ref{sec:WBLC_setting}$, for locomotion we define a foot task and a body posture task, $\mathcal{X}_{\textrm{Mercury}} = \{ \ddot{\mathbf{x}}_{foot}, \ddot{\mathbf{x}}_{body} \}$,
where $\ddot{\mathbf{x}}_{foot}$ and $\ddot{\mathbf{x}}_{body}$ are specifications for the foot and body tasks. The height, roll, and pitch of the body are controlled as constant values, respectively. Since DRACO includes hip joints on both left and right side legs, we additionally formulate a hip configuration task for both hip joints of DRACO in addition to Mercury's tasks, $\mathcal{X}_{\textrm{DRACO}} = \mathcal{X}_{\textrm{Mercury}} \cup \{ \ddot{\mathbf{x}}_{hip} \}$. The body task of DRACO controls its body height, roll, pitch and yaw orientation. As shown in Fig \ref{fig:sim_multi_robots} (a) and (b), the simulation results of Mercury and DRACO demonstrate that both robot simulations are able to perform dynamic walking without much algorithmic modifications. The parameters of the planner are set to $t' = [0.2, \: 0.2]$ and $\kappa = [0.16, \: 0.16]$   

Unlike the above two robots, Atlas and Valkyrie are full-body humanoid robots and their ankle joints are actuated so that we modify the task sets and constraints to test our algorithm using simulations. We modify the inequality constraint for the contact wrench cone to surface contacts in (\ref{eq:wblc_cone_const}). For these full-body humanoid robots, the height of the pelvis, which corresponds to the floating base, is constantly controlled in the same way than Mercury and DRACO. We define the orientation tasks for the pelvis and torso. Also, a task for controlling foot orientation is introduced for stable contact on the feet. Based on the defined tasks, the task set of Atlas is designed to be $\mathcal{X}_{\textrm{Atlas}} = \{\ddot{\mathbf{x}}_{foot},\ddot{\mathbf{x}}_{pelvis}, \ddot{\mathbf{x}}_{torso}, \ddot{\mathbf{x}}_{jpos} \}$ where $\ddot{\mathbf{x}}_{jpos}$ represents a task for controlling the entire joint positions of the robot. As shown in the simulation, Atlas is able to perform dynamic walking similarly to Mercury and DRACO without modifying our algorithms as shown in Fig. \ref{fig:sim_multi_robots} (c). 

We define additional tasks for controlling the left hand and the head orientations to demonstrate locomanipuation capabilities on Valkyrie, $\mathcal{X}_{\textrm{Valkyrie}} = \mathcal{X}_{\textrm{Atlas}} \cup \{ \ddot{\mathbf{x}}_{hand}, \ddot{\mathbf{x}}_{head} \}$. The simulation result shows that our algorithm can accomplish the desired locomanipulation behavior as shown in Fig.~\ref{fig:sim_multi_robots} (d). These four simulations show that our algorithm is applicable to various biped humanoids. 

\section{Conclusions}
\label{sec:conclusion}

We demonstrated here robust dynamic walking of various biped robots, including one with no ankle actuation, using a novel locomotion-control scheme consisting of two components dubbed WBLC controller and TVR planner. The algorithmic generality has been verified on hardware with the bipeds Mercury and DRACO and in simulation with other humanoids such as Valkryie and Atlas. We have performed an uncertainty analysis of the TVR planner and found maximum allowable errors for our state estimator and controllers, which enabled us to significantly redesign and rebuild the Mercury robot and tune the controllers and estimators. By integrating a high-performance whole-body feedback controller, WBLC, a robust locomotion planner, TVR, and a reliable state estimator, our passive-ankle biped robot and lower body humanoid robot accomplish unsupported dynamic locomotion robust to impact disturbances and rough terrains.

In devising our control scheme, we have experimented with a variety of whole-body control formulations and feedback controllers. We compared different WBC operational task specifications such as foot position vs leg joint position control, base vs CoM position control, having vs not having task priorities, etc. In the low-level controller we also experimented with torque feedback with disturbance observers, joint vs motor position feedback, and joint position control with and without feedforward torques. The methodology presented here is our best performing controller after system-level integration and exhaustive testing. 

With our new biped, DRACO, we have explored initial dynamic locomotion. In the future, we will explore more versatile locomotion behaviors such as turning and walking in a cluttered environment. In the case of Mercury, we could not change the robot's heading because of the lack of yaw directional actuation. With simple additions to the current TVR planner, we will be able to test turning of DRACO since the robot has hip yaw actuation. In addition, we will conduct robustness tests in a more complex way by exploring a cluttered environment involving contacts with many objects including human crowds.

\section*{ACKNOWLEDGMENTS}
The authors would like to thank the members of the Human Centered Robotics Laboratory at The University of Texas at Austin and the members of Apptronik for their support. This work was supported by the Office of Naval Research, ONR grant \#N000141512507, NSF grant \#1724360, and a NASA Space Technology Research Fellowship (NSTRF) grant \#NNX15AQ42H.

\bibliographystyle{SageH}
\bibliography{ijrr}

\section*{Appendix 1: Index to multimedia extensions}
A video showing experiment and simulation results from Section.~\ref{sec:experiment} is included.
\begin{table}[h]
\centering
\begin{tabular}{>{}m{0.2\columnwidth} %
                >{}m{0.22\columnwidth} %
                >{}m{0.4\columnwidth} %
                @{}m{0pt}@{}}
\specialrule{1.5pt}{1pt}{1pt}
{Extension}
& {Media type}
& {Description}
&\\[1mm] 
\hline
\hline 
1 & Video & A video of experimental and simulation results from Section.~\ref{sec:experiment}
& 
\\[1.5mm]
\hline
\end{tabular}
\caption{Table of Multimedia Extensions}
\label{tb:multi_media}
\vspace{-1.5mm}
\end{table}

\section*{Appendix 2: Difference Among Footstep Planners}
\label{sec:planner_diff}

The footstep planners proposed in  \cite{Raibert:1984ej}, \cite{Koolen:2012cta}, \cite{Rezazadeh:vk}, and ours are sharing the idea that footsteps are chosen based on a weighted sum of the CoM state, which is formulated by the general equation, 
\begin{equation}\label{eq:footstep_general_eq}
    p = k_p x + k_d \dot{x},
\end{equation}
where $p$, $x$, $\dot{x}$, $k_p$ and $k_d$ are foot landing positions, sensed CoM position and velocity, and weights (or gains) for position and velocity feedback, respectively. This equation can be slightly varied by including desired position or velocity terms, but the basic idea does not change because of these additions.

\cite{Raibert:1984ej} introduced long ago an unsupported hopping robot and its control method. For hopping control, the foot placement is decided by the equation,
\begin{equation}
    p = \frac{T_{st}}{2}\dot{x} + K \left(\dot{x} - \dot{x}_d \right),
\end{equation}
where $T_{st}$ is the duration of the stance phase. Since the desired velocity, $\dot{x}_d$, is defined by the equation, $\dot{x}_d = -K_p x - K_v \dot{x}$ in the paper, the resulting equation for foot placement becomes 
\begin{equation}
    p = KK_p x +\left( K(K_v+1) + \frac{T_{st}}{2} \right)\dot{x},
\end{equation}
which has similar form to Eq.~\eqref{eq:footstep_general_eq} with $k_p = K K_p$ and $k_v = K(K_v+1) + T_{st}/2$.

\cite{Koolen:2012cta} proposed a method called capture point (CP), which computes the foot's center of pressure (CoP) to drive the CoM velocity to zero at the CoP location given the current CoM state. Using the LIP model, CP is defined by the equation, 
\begin{equation}
    cp = x + \sqrt{\frac{h}{g}}\dot{x},    
\end{equation}
where $g$ and $h$ are the gravitational acceleration and the CoM height, respectively. This can be expressed using the form of Eq.~\eqref{eq:footstep_general_eq} by plugging 1 into $k_p$ and $\sqrt{h/g}$ into $k_d$. As we mentioned above, given CoM state, a robot will stop and stay on top of the capture point if the robot maintains its CoP on the CP. Differently, our TVR planner finds a foot placement location such that it reverses the CoM velocity before the CoM reaches that location.

In \cite{Rezazadeh:vk}, the step location for in-place walking is provided by the equation,
\begin{equation}
    p = K_P \dot{x} + K_D (\dot{x} - \dot{x}_{n-1}) + K_I x,
\end{equation}
which has a slightly different form than the previous controllers because of the term $\dot{x}_{n-1}$ representing the CoM velocity at the previous step. However, for in-place walking, the effect of the velocity error between the current and previous step is not very significant. Therefore, we can regard the above equation as one variation of Eq.~\eqref{eq:footstep_general_eq}. In the same vein, our TVR planner, as presented in Eq.~\eqref{eq:tvr_step_eq}, is also a variation of Eq.~\eqref{eq:footstep_general_eq}, with weights, $k_p=(1+\kappa)$ and $k_d = w^{-1}\coth(\omega t')$. 

In conclusion, various locomotion planners can be casted using variations of Eq.~\eqref{eq:footstep_general_eq}. The resulting behaviors vary depending on the chosen weights, e.g. CP makes a robot stop while ours makes the robot reverse its direction of motion. The benefit of our TVR planner over the others is that we provide an intuitive method and analysis to help with feedback gain selection. Our planner parameter $t'$ must be close to half of the designated swing time and additional tuning for asymptotic stability is possible by checking the eigenvalues of the matrix, $\bm{A}+\bm{B}\bm{K}$ in Eq.~\eqref{eq:tvr_closed_form}.

\section*{Appendix 3: Definition of $g(\bm{\zeta}^{\top}\bm{P}\bm{\zeta})$}

\begin{equation}
    g(\bm{\zeta}^{\top}\bm{P}\bm{\zeta}) = \delta_M^2 D + 2\delta_M \eta_M E + \eta_M^2 F,
\end{equation}
where 
\begin{align}
    &D = ||(\bm{A}_c\bm{B}\bm{K})^{\top}\bm{P}\bm{A}_c\bm{B}\bm{K}||+ 2||(\bm{A}_c\bm{B}\bm{K})^{\top}\bm{P}\bm{B}\bm{K}|| \nonumber \\
    &\quad \quad + ||(\bm{B}\bm{K})^{\top}\bm{P}\bm{B}\bm{K}||, \\[2mm]
    &E = ||(\bm{A}_c\bm{B}\bm{K})^{\top}\bm{P}\bm{A}_c\bm{B}|| + ||(\bm{A}_c\bm{B}\bm{K})^{\top}\bm{P}\bm{B}|| \nonumber \\
    &\quad \quad +  ||(\bm{A}_c\bm{B})^{\top}\bm{P}\bm{B}\bm{K}|| +||\bm{B}^{\top}\bm{P}\bm{B}\bm{K}||, \\[2mm]
    &F = ||(\bm{A}_c\bm{B})^{\top}\bm{P}\bm{A}_c\bm{B}|| + 2||\bm{B}^{\top}\bm{P}\bm{A}_c\bm{B}|| + ||\bm{B}^{\top}\bm{P}\bm{B}||.
\end{align}

\end{document}